\documentclass[conference]{IEEEtran}
\usepackage{amsmath}
\usepackage{array}
\usepackage{booktabs}
\usepackage{cite}
\usepackage{fancyhdr}
\usepackage{graphicx}
\usepackage{microtype}
\usepackage{url}
\usepackage{tikz}
\usetikzlibrary{arrows.meta,calc,positioning,shapes.misc}
\usepackage{pgfplots}
\pgfplotsset{compat=1.18}
\usepgfplotslibrary{fillbetween}

\newcolumntype{P}[1]{>{\raggedright\arraybackslash}p{#1}}

\title{Tracing Attacker Diversity Reduction in Adaptive LLM Red Teaming}

\author{
\IEEEauthorblockN{Chris Tava}
\IEEEauthorblockA{christava@microsoft.com}
}

\begin{document}

\maketitle

\begin{abstract}
Reinforcement learning can increase the success of automated red teaming but it also narrows the attack vector. This study asks where in the attacker--defender--judge learning loop does the narrowing begin and which mechanisms lead to a narrowing attack. Six mechanisms (defender adaptation, success edge, reward reinforcement, row cap, update size, and measurement) are tested across three experiment arms (co-trained, frozen defender, half learning rate) for twenty-four iterations. A frozen defender narrows attack diversity retention (ADR) earlier 0.496 relative to co-trained 0.891 at iteration twelve. Defender updates delay narrowing but dont prevent it. Successful attacks ("winners") have narrower attack diversity than the full set of generated attacks. Attacks closest to previous winners succeed 10–15 percentage points more often than those closest to previous failures. This winner-neighbor advantage holds after a defender update and for a fixed defender too. Attacks with a common opening have a common-opening advantage of 5--16 percentage points. The training row cap and update size account for little impact on ADR, and attack length accounts for only part of the ADR decrease. Narrowing appears to originate from the success-only reward through the group-relative training advantage to the attacker update, with a persistent success edge setting the direction.
\end{abstract}

\section{Introduction}

Red teaming with a trained attacker model extends hand-written test cases to automated scenarios \cite{perez2022redteaming,zou2023universal}, and reinforcement learning allows the attacker to adjust its strategy in response to the defender. Co-training an attacker with group-relative policy optimization (GRPO) \cite{shao2024deepseekmath} and a defender with direct preference optimization (DPO) \cite{rafailov2023dpo} has been reported to reduce attacker diversity \cite{bullwinkel2026advgrpo}. A narrow attack set exposes the defender to a less diverse set of attacks, reducing the coverage that red teaming is meant to provide.

A previous study reproduces attacker diversity reduction with the same co-training setup as the co-trained arm in this study \cite{tava2026adrrepro}. That paper documented the reduction in attacker diversity but did not investigate its cause. Several points in the attacker--defender--judge learning loop could contribute to narrowing: defender adaptation, judge scoring, training advantage assignment, training-row selection, attacker updates and the diversity measure itself.

This paper asks \emph{where diversity reduction originates in the attacker--defender--judge learning loop and based on the results of the experiments, which of six mechanisms is likely responsible.} Section~\ref{sec:loop} describes the mechanisms and their expected outcomes. Results are captured in three arms---co-training, a frozen defender and half the attacker learning rate---that share a seed and results are recorded using records of attacks, winners and training rows from each iteration, plus 72 updates retained from three earlier runs.

The paper locates where narrowing appears in the learning loop and evaluates alternative explanations, without claiming to establish a single cause. The evidence is limited to one seed per arm, includes several exploratory analyses, and uses one judge and one encoder. Section~\ref{sec:limits} describes these limits, which bound the conclusion.

\textbf{Contributions.}
\begin{enumerate}
    \item \emph{Decomposition of the learning loop into six mechanisms,} each with observations that can be reviewed against the recorded outcomes (Section~\ref{sec:loop}).
    \item \emph{Shared-seed arms and a mechanism record:} 24 iterations of co-training, a frozen defender and half the attacker learning rate from one seed, with per-iteration records of attacks, winners, selected rows, openings, length and update size (Sections~\ref{sec:design} and~\ref{sec:results}).
    \item \emph{Mechanism-by-mechanism assessment:} Narrowing continues even with a frozen defender. In every retained update comparison, winners are narrower than all valid attacks, starting with the first iteration. Every arm shows both a winner-neighbor advantage and a common-opening advantage. The row cap and update transients contribute little to attacker diversity reduction, while attack length explains part of the ADR decline (Section~\ref{sec:discussion}).
\end{enumerate}

\section{The Learning Loop and Mechanisms}
\label{sec:loop}

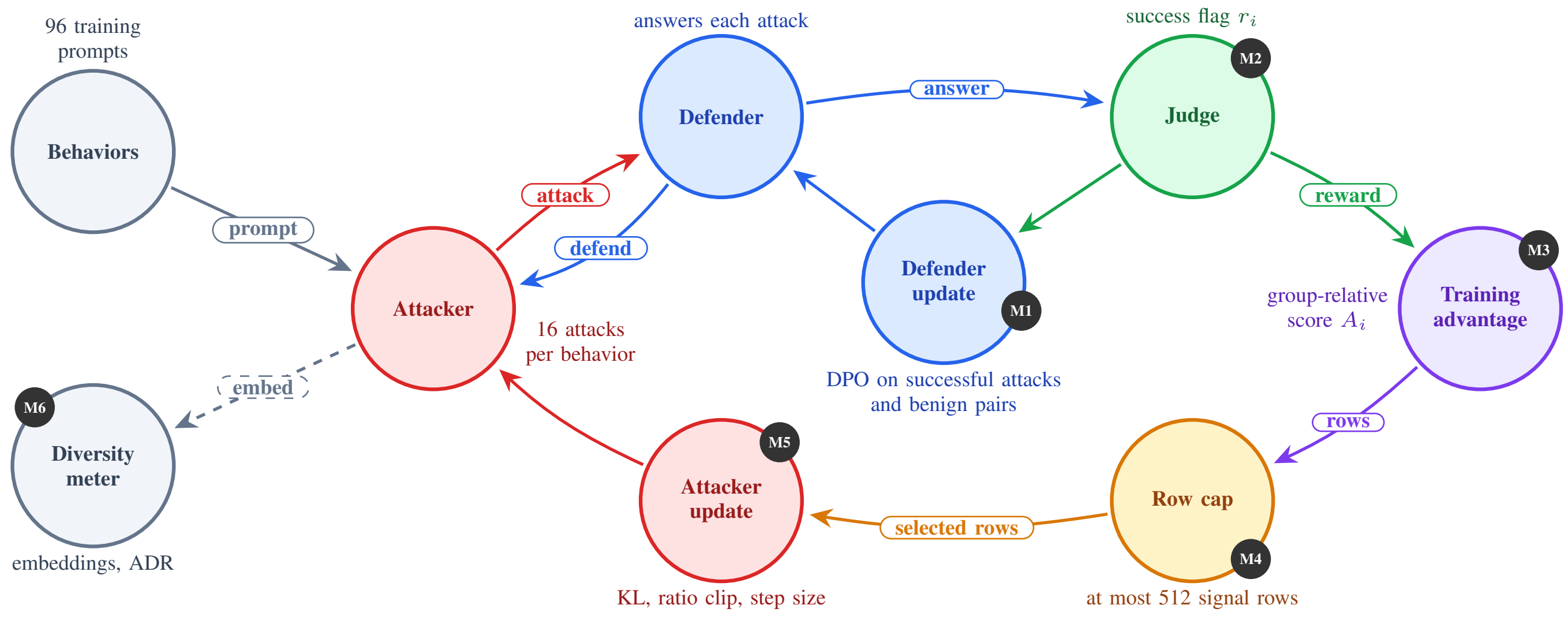
\begin{figure*}[t]
\centering
\definecolor{dataS}{HTML}{64748B}\definecolor{dataF}{HTML}{F1F5F9}\definecolor{dataT}{HTML}{334155}
\definecolor{atkS}{HTML}{DC2626}\definecolor{atkF}{HTML}{FEE2E2}\definecolor{atkT}{HTML}{991B1B}
\definecolor{defS}{HTML}{2563EB}\definecolor{defF}{HTML}{DBEAFE}\definecolor{defT}{HTML}{1E40AF}
\definecolor{judS}{HTML}{16A34A}\definecolor{judF}{HTML}{DCFCE7}\definecolor{judT}{HTML}{166534}
\definecolor{advS}{HTML}{7C3AED}\definecolor{advF}{HTML}{EDE9FE}\definecolor{advT}{HTML}{5B21B6}
\definecolor{rowS}{HTML}{D97706}\definecolor{rowF}{HTML}{FEF3C7}\definecolor{rowT}{HTML}{92400E}
\begin{tikzpicture}[
    >=Stealth,
    comp/.style={circle, draw, line width=1.2pt, minimum size=1.85cm, inner sep=0pt, align=center, font=\scriptsize\bfseries},
    sub/.style={font=\scriptsize, align=center, inner sep=1pt},
    link/.style={line width=1pt, -{Stealth[length=2.6mm,width=2.2mm]}, shorten <=0.4mm, shorten >=0.8mm},
    pill/.style={fill=white, rounded rectangle, rounded rectangle arc length=180, inner xsep=4pt, inner ysep=1.2pt, line width=0.5pt, font=\scriptsize\bfseries, midway},
    tag/.style={circle, fill=black!80, text=white, inner sep=0.8pt, minimum size=0.46cm, font=\tiny\bfseries}]
  \node[comp, fill=dataF, draw=dataS, text=dataT] (beh) at (-3.9,1.8)  {Behaviors};
  \node[comp, fill=atkF,  draw=atkS,  text=atkT]  (atk) at (0,0)       {Attacker};
  \node[comp, fill=defF,  draw=defS,  text=defT]  (def) at (3.3,2.2)   {Defender};
  \node[comp, fill=judF,  draw=judS,  text=judT]  (jud) at (8.7,2.2)   {Judge};
  \node[comp, fill=advF,  draw=advS,  text=advT]  (adv) at (12.0,0)    {Training\\advantage};
  \node[comp, fill=rowF,  draw=rowS,  text=rowT]  (row) at (8.7,-2.2)  {Row cap};
  \node[comp, fill=atkF,  draw=atkS,  text=atkT]  (upd) at (3.3,-2.2)  {Attacker\\update};
  \node[comp, fill=defF,  draw=defS,  text=defT]  (dup) at (5.85,0.3)  {Defender\\update};
  \node[comp, fill=dataF, draw=dataS, text=dataT] (div) at (-3.9,-1.8) {Diversity\\meter};
  \node[sub, text=dataT, above=1pt of beh] {96 training\\prompts};
  \node[sub, text=atkT,  right=2pt of atk, yshift=-4mm] {16 attacks\\per behavior};
  \node[sub, text=defT,  above=1pt of def] {answers each attack};
  \node[sub, text=judT,  above=1pt of jud] {success flag $r_i$};
  \node[sub, text=advT,  left=2pt of adv]  {group-relative\\score $A_i$};
  \node[sub, text=rowT,  below=1pt of row] {at most 512 signal rows};
  \node[sub, text=atkT,  below=1pt of upd] {KL, ratio clip, step size};
  \node[sub, text=defT,  below=1pt of dup] {DPO on successful attacks\\and benign pairs};
  \node[sub, text=dataT, below=1pt of div] {embeddings, ADR};
  \draw[link, dataS] (beh) -- node[pill, draw=dataS, text=dataS] {prompt} (atk);
  \draw[link, atkS] (atk) to[bend left=9] node[pill, draw=atkS, text=atkS] {attack} (def);
  \draw[link, defS] (def) to[bend left=18] node[pill, draw=defS, text=defS] {defend} (atk);
  \draw[link, defS] (def) to[bend left=9] node[pill, draw=defS, text=defS] {answer} (jud);
  \draw[link, judS] (jud) to[bend left=9] node[pill, draw=judS, text=judS] {reward} (adv);
  \draw[link, advS] (adv) to[bend left=9] node[pill, draw=advS, text=advS] {rows} (row);
  \draw[link, rowS] (row) to[bend left=9] node[pill, draw=rowS, text=rowS] {selected rows} (upd);
  \draw[link, atkS] (upd) to[bend left=9] (atk);
  \draw[link, judS] (jud) -- (dup);
  \draw[link, defS] (dup) -- (def);
  \draw[link, dataS, dashed] (atk) -- node[pill, draw=dataS, text=dataS] {embed} (div);
  \node[tag] at (jud.45) {M2};
  \node[tag] at (adv.45) {M3};
  \node[tag] at (row.-45) {M4};
  \node[tag] at (upd.45) {M5};
  \node[tag] at (dup.-20) {M1};
  \node[tag] at (div.135) {M6};
\end{tikzpicture}
\caption{One attacker iteration follows the outer loop: the attacker generates attacks, the defender responds, and the judge scores each attack. The defend arrow represents the defender response, which is sent to the judge for scoring. Within each behavior, the scores determine group-relative training advantages that weight attacks for training. Up to 512 signal rows are selected for the attacker update. In arms with an updated defender, successful attacks also inform one defender update after every second attacker iteration. The diversity meter measures attack diversity. Labels M1--M6 mark the mechanisms in Table~\ref{tab:mechanisms}.}
\label{fig:loop}
\end{figure*}

Figure~\ref{fig:loop} shows one pass through the learning loop. In each iteration the attacker writes 16 attacks for each of 96 training behaviors. The defender answers every attack, and the judge returns a success flag, which is the entire reward. Within the 16 attacks of one behavior, the flag becomes a group-relative (GRPO) training advantage. Attacks with a nonzero training advantage form the signal rows. A seed-determined uniform selection of up to 512 signal rows is used for one attacker update, after which the updated attacker writes the next iteration. In the arms with an updated defender, successful attacks also yield preference pairs for one defender update after every second attacker iteration. A diversity meter creates embeddings for valid attacks and reports attack diversity retention (ADR), defined in Section~\ref{sec:measures}.

This study considers six mechanisms that could account for a decline in ADR, each acting at a distinct point in the learning loop. \emph{Defender adaptation} (M1): an updated defender learns to refuse attacks that bypassed its earlier version, leaving the attacker with a narrower set of attacks that still work. \emph{Success edge} (M2): under a fixed judge and base defender, some attacks succeed more often than others for reasons unrelated to training, and any learner that follows success concentrates on them. \emph{Reward reinforcement} (M3): the success-only training advantage could raise the probability of successful attacks, making the next iteration more likely to produce them and the cycle compound. \emph{Row cap} (M4): a seed-determined uniform selection of at most 512 rows could over- or under-weight some attacks. \emph{Update size} (M5): a few very large updates could push the attacker into a narrow region from which later updates return slowly. \emph{Measurement} (M6): ADR is a ratio of embedding distances and can decrease with attack length, embedding geometry or the coarse label mix while the variety of strategies holds.

Table~\ref{tab:mechanisms} states the expected outcome for each mechanism if it were the main origin of attacker diversity reduction. Section~\ref{sec:results} compares those expectations with the records, and Section~\ref{sec:discussion} gives the standing of each mechanism. M2 and M3 are complementary: M2 supplies a direction and M3 an amplifier. Although the records are consistent with both, they cannot establish whether reward reinforcement amplifies the success edge. A controlled intervention on the reward signal would test that link (Section~\ref{sec:discussion}).

\begin{table*}[t]
\centering
\caption{Mechanisms, their place in the loop and the outcome expected if each were the main origin of drop in attack diversity.}
\label{tab:mechanisms}
\small
\begin{tabular}{@{}P{0.17\textwidth}P{0.14\textwidth}P{0.37\textwidth}P{0.24\textwidth}@{}}
\toprule
Mechanism & Place in the loop & Expected outcome if it were the main origin & Evidence considered \\
\midrule
M1 Defender adaptation & Defender update & Narrowing is weaker or absent when the defender stays fixed. Onset follows defender updates. The defender fits the successful attacks. & Frozen-defender arm, defender fit, winner-neighbor advantage across defender updates \\
\addlinespace
M2 Success edge & Judge and base defender & Some attacks succeed more often even when the defender stays fixed. Attacks similar to earlier winners also succeed more often. & Winner-neighbor advantage, common-opening advantage \\
\addlinespace
M3 Reward reinforcement & Training advantage to attacker update & Winners are narrower than the distribution of generated attacks. Attack-family shares in the next iteration follow enrichment among winners. The training objective has no explicit diversity reward. & Winner dispersion ratio, family and opening shares, reward definition \\
\addlinespace
M4 Row cap & Row selection & Selected rows are more concentrated, or enriched in one attack category, than all signal rows & Selected rows against all rows \\
\addlinespace
M5 Update size & Attacker update & Narrowing tracks the size of transient updates. A halved learning rate prevents narrowing. & Half-step arm, mean KL and gradient norm per update \\
\addlinespace
M6 Measurement & Diversity meter & The diversity reduction follows attack length or coarse label count. Length-adjusted and wording-based readings stay near 1. & ADR after length adjustment, word 3-gram distance, family mix \\
\bottomrule
\end{tabular}
\end{table*}

\section{Design}
\label{sec:design}

\subsection{Models, Data and the Attacker Update}

Table~\ref{tab:config} lists the models and settings shared by all arms. The attacker is Qwen2.5-14B-Instruct \cite{qwen2024qwen25} with a low-rank adapter (LoRA) \cite{hu2022lora}. The defender is Phi-4-mini-instruct \cite{microsoft2025phi4minireport} with a LoRA adapter trained by direct preference optimization (DPO) \cite{rafailov2023dpo}. The HarmBench Llama-2-13b classifier \cite{mazeika2024harmbench} serves as judge, and bge-large-en-v1.5 embeddings \cite{xiao2024cpack} define attack distance. A master seed splits the AdvBench behaviors \cite{zou2023universal}, with the chemical and biological category excluded, into training, development (26 behaviors) and test behaviors. A seeded panel of 96 training behaviors supplies the rollouts of every arm. Benign prompts drawn from the Alpaca instructions \cite{taori2023alpaca} guard the defender against over-refusal. The attacker prompt carries the behavior and the context of an authorized safety evaluation. Strategy descriptions and examples stay out of the prompt, so the attacker chooses the strategy.

\begin{table}[t]
\centering
\caption{Models and settings shared by all arms. The half-step arm halves the attacker peak learning rate. The frozen-defender arm omits the defender update.}
\label{tab:config}
\small
\begin{tabular}{@{}P{0.27\columnwidth}P{0.65\columnwidth}@{}}
\toprule
Setting & Value \\
\midrule
Attacker & Qwen2.5-14B-Instruct \cite{qwen2024qwen25}, LoRA \cite{hu2022lora} rank 16, $\alpha=32$ \\
Defender & Phi-4-mini-instruct \cite{microsoft2025phi4minireport}, LoRA rank 16, $\alpha=32$, DPO \cite{rafailov2023dpo} \\
Judge & HarmBench-Llama-2-13b classifier \cite{mazeika2024harmbench} \\
Embedder & bge-large-en-v1.5 \cite{xiao2024cpack} \\
\midrule
Rollout & 16 attacks for each of 96 training behaviors, temperature 1.0, at most 300 new tokens \\
Reward & success flag, with quality and novelty weights set to 0 \\
Attacker update & AdamW \cite{loshchilov2019adamw}, peak learning rate $2\times10^{-4}$ ($1\times10^{-4}$ at half step), four-step warm-up, gradient-norm clip 1.0, eight rows per step, KL weight 0.005, ratio clip 0.2, constant loss normalizer 300 \\
Update input & Skipped below 64 signal rows. Otherwise, up to 512 rows are selected uniformly using the seed. \\
Defender update & DPO with learning rate $2\times10^{-5}$, one epoch and $\beta=0.1$. At most 64 attack pairs and 32 benign retention pairs yield about 12 optimizer steps. Decoding temperature is 0, with at most 600 new tokens. \\
Schedule & A round is two attacker iterations, then one defender update. The schedule runs for 12 rounds. \\
\midrule
Development banks & 26 AdvBench behaviors, 8 attacks each, temperature 1.0, scored against the untrained defender after every iteration \\
Over-refusal guard & 56 benign Alpaca prompts \\
Intervals & 2,000 paired behavior-bootstrap draws \cite{efron1979bootstrap} \\
\bottomrule
\end{tabular}
\end{table}

For the $G=16$ attacks written for one behavior, the reward and the group-relative training advantage are
\begin{align}
    r_i &= \mathbf{1}[\text{success}_i],\\
    A_i &= \frac{r_i-\bar r}{\sigma_r+\eta},
\end{align}
Here, $i$ indexes an attack in the group, and $r_i$ is its success flag. The group-relative training advantage $A_i$ measures how the reward for the attack compares with the group mean $\bar r$, scaled by the reward standard deviation for the group $\sigma_r$. The constant $\eta=10^{-6}$ prevents division by zero. In a group with both successes and failures, successful attacks have positive training advantage and failures have negative training advantage. Groups with uniform rewards have zero training advantage. A \emph{winner} is a valid attack with $r_i=1$, and a \emph{signal row} is an attack with nonzero training advantage. The reward is the success flag alone. Quality and novelty weights remained zero throughout every arm.

For attack $i$, let $x$ be its behavior prompt and $y=(y_1,\ldots,y_{|y|})$ its generated tokens. The prefix $y_{<t}$ contains the tokens before position $t$. Let $\pi_c$ be the current attacker policy being updated, $\pi_\mathrm{old}$ the attacker policy at the start of the update and $\pi_b$ the base model policy with the adapter disabled. Define
\begin{equation}
    \rho_t=\frac{\pi_c(y_t\mid x,y_{<t})}{\pi_\mathrm{old}(y_t\mid x,y_{<t})},\qquad
    \delta_t=\log\frac{\pi_b(y_t\mid x,y_{<t})}{\pi_c(y_t\mid x,y_{<t})}.
\end{equation}
The ratio $\rho_t$ is the probability assigned to next token $y_t$ by $\pi_c$, divided by the probability assigned to the same token by $\pi_\mathrm{old}$, with prompt $x$ and prefix $y_{<t}$ held fixed. It therefore compares conditional next-token probabilities, not probabilities of the complete attack sequence. A value above 1 means the current policy assigns a higher probability to that token in this context. The log ratio $\delta_t$ compares the base and current policy probabilities for the same token and context. The group-relative training advantage $A_i$ assigned to attack $i$ weights each of its tokens. For one signal row, the loss $\mathcal{L}$ combines a clipped policy objective, based on $\rho_t$, and a Kullback--Leibler (KL) penalty, based on $\delta_t$:
\begin{equation}
\begin{split}
\mathcal{L} = \frac{1}{C}\sum_{t=1}^{|y|}\Big[&-\underbrace{\min\big(\rho_t A_i,\ \mathrm{clip}(\rho_t,1-\varepsilon,1+\varepsilon)\,A_i\big)}_{\text{clipped policy objective}}\\
 &+\underbrace{\beta\left(e^{\delta_t}-\delta_t-1\right)}_{\text{KL penalty}}\Big].
\end{split}
\label{eq:loss}
\end{equation}
The clipping parameter $\varepsilon=0.2$ sets the interval $[1-\varepsilon,1+\varepsilon]=[0.8,1.2]$ \cite{schulman2017ppo}. The clip function replaces a ratio outside this interval with the nearest endpoint and leaves a ratio inside it unchanged. The clipped policy objective multiplies both the original and clipped ratios by the training advantage, then takes the smaller result. The leading minus sign means that increasing this objective decreases its contribution to the loss. For $A_i>0$, the policy term favors increasing the probability of tokens in a successful attack relative to $\pi_\mathrm{old}$. For $A_i<0$, it favors decreasing the probability of tokens in an unsuccessful attack. When $\rho_t>1.2$ for a successful attack, or $\rho_t<0.8$ for an unsuccessful attack, the objective no longer favors further change in that direction. Clipping does not force $\rho_t$ to stay inside the interval or remove the KL penalty.

The KL penalty compares the current policy $\pi_c$ with the base policy $\pi_b$, not with $\pi_\mathrm{old}$. Its weight $\beta=0.005$ controls its contribution to the loss. The unweighted expression $e^{\delta_t}-\delta_t-1$ is nonnegative and is zero when the current and base policies assign the same probability to the token in the given context. Minimizing this term favors bringing the current policy probability for the generated token closer to the base-model probability. For a fixed prompt and prefix, averaging this expression over all possible next tokens, weighted by their probabilities under $\pi_c$, gives the KL divergence from $\pi_c$ to $\pi_b$ \cite{shao2024deepseekmath}. The update evaluates the penalty on selected tokens generated before the update, so this per-token estimate is not the exact divergence between the full next-token distributions. The constant normalizer $C=300$, the generation limit, gives each token the same scaling regardless of attack length \cite{liu2025drgrpo}. Uniform-reward groups contribute zero rows \cite{yu2025dapo}, and the learning rate warms up over the first four optimizer steps of every update.

\subsection{Arms and Schedule}

Three arms start from the same untrained models and share a seed, the 96-behavior training panel, the 26-behavior development panel and every setting of Table~\ref{tab:config} except the two interventions.
\begin{itemize}
    \item \emph{Co-trained:} The attacker peak learning rate is $2\times10^{-4}$. The defender is updated by DPO after every second attacker iteration.
    \item \emph{Frozen defender:} The defender stays at the untrained model throughout, so the arm has zero defender updates.
    \item \emph{Half step:} The attacker peak learning rate is $1\times10^{-4}$ in every update. The defender is updated as in the co-trained arm.
\end{itemize}
Each round consists of two attacker iterations, followed by a defender update in the co-trained and half-step arms. The frozen-defender arm has no defender update. Each arm has one run. All arms use the same seed and the same procedures for selecting inputs and initializing random-number generators. Their generated texts differ. Every contrast with the co-trained arm is a total effect of the intervention, including the changes in success, signal rows and attacker policy that follow from it. The pre-specified analysis covered the first 12 iterations (6 rounds). The arms were then continued from saved states to 24 iterations (12 rounds) with unchanged settings. Each iteration produced an update, giving 24 attacker updates per arm and 12, 0 and 12 defender updates, respectively.

\subsection{Retained Runs}

Three earlier co-training runs reported in the previous study \cite{tava2026adrrepro}, provide relevant evidence. Each run used a distinct seed and completed 24 attacker updates and 12 defender updates. Across the 72 attacker updates, the records contain 110,592 attack slots. Of these, 110,312 contained valid attacks and 28,012 were recorded as successful. The records also show 82,240 signal rows, of which 36,656 were selected for attacker training across 4,582 optimizer steps. The saved records were analyzed again for this study. The attacks, defender responses and judge labels used in this paper were recorded during the runs, before any of the analyses reported here were conducted. All 67 updates that used the row cap reproduce the recorded row selection from the seed.

\subsection{Measures}
\label{sec:measures}

\textbf{Attack diversity retention.} A development bank is a set of eight attacks generated for each of the 26 development behaviors. The untrained bank, shared by all arms, contains attacks generated before the first attacker update (checkpoint 0). A new bank is generated after each attacker iteration. The endpoint bank is the one generated at the final iteration being measured (iteration 12 or 24 for cumulative change). A valid attack has at least 20 and fewer than 1,500 characters and passes an opening-text refusal filter. For behavior $b$ at checkpoint $u$, let $d_b(u)$ be the mean cosine distance between all pairs of embeddings of valid attacks written for $b$. For the set $\mathcal{B}$ of behaviors with at least two valid attacks in both the untrained bank and the bank at checkpoint $u$,
\begin{equation}
    \mathrm{ADR}(u)=\frac{\sum_{b\in\mathcal{B}} d_b(u)}{\sum_{b\in\mathcal{B}} d_b(0)},
\end{equation}
Values below 1 mean narrower within-behavior spread. Intervals are calculated from 2,000 bootstrap draws of behaviors, with each behavior kept paired across the banks being compared. They describe uncertainty across the behavior panel for one seed per arm. Each cumulative-change estimate uses 25 of the 26 development behaviors, with at least two valid attacks for each behavior in both the untrained and endpoint banks. To estimate the half-step trend over iterations 3 to 12, ADR is calculated at each iteration from the same 24 behaviors. Each behavior has at least two valid attacks in the untrained bank and after every iteration in that period. A straight line is fitted to the ten ADR values to estimate the change per iteration. The comparison with the co-trained arm uses the same behaviors, which meet these requirements in both arms.

\textbf{Primary readings.} Before the first iteration, the study defined two measurements for each arm. The first is the change in ADR from the untrained attacker to iteration 12, calculated as $\mathrm{ADR}(12)-1$. The second is the rate of change in ADR per iteration from iteration 3 to iteration 12, estimated by fitting a straight line by least squares to the ten ADR values. For both measurements, differences from the co-trained arm were also calculated for the frozen-defender and half-step arms. After the study was extended to 24 iterations, the change in ADR from the untrained attacker to iteration 24 and the slope over iterations 13 to 24 were calculated as well. These later readings and their comparisons with the co-trained arm are exploratory. The value 0.75, a loss of one quarter of the within-behavior spread, represents attacker diversity reductions and serves as a reference line. The continuous readings carry the conclusions.

\textbf{Success.} ASR@1 is the share of development behaviors with a successful first attack, ASR@8 the share with at least one success among eight attacks, and the success fraction the share of all 208 attacks that succeed. The development banks are scored against the untrained defender in every arm, so arms compare directly. Training rollouts are scored against the current defender of the arm.

\textbf{Mechanism record.} Every iteration stores, for each of the 1,536 attack slots, the behavior, the validity, the success flag, the training advantage, whether the attack was selected for the attacker update, the embedding, an automated strategy-family label and the first four words. The readings below use this record alone.
\begin{itemize}
    \item \emph{Winner dispersion ratio} (M3): over behaviors with at least two winners, the mean pairwise embedding distance among winners divided by the mean pairwise distance among all valid attacks of the same behaviors. Values below 1 mean that winners are narrower than the pool.
    \item \emph{Winner-neighbor advantage} (M1, M2): an attack has a winner neighbor when its nearest neighbor among attacks written for the same behavior at the previous iteration was a winner. For each behavior at each iteration, the success rate of attacks with a winner neighbor is compared with the rate for other attacks. These differences are combined across behaviors and iterations using Mantel--Haenszel weights \cite{greenland1985mh}. The uncertainty intervals are calculated by repeatedly choosing groups of behaviors from the recorded set, allowing the same behavior to appear more than once, and recalculating the winner-neighbor advantage for each group. In the arms with defender updates, iterations 3, 5, 7, 9 and 11 each follow an update. The defender contrast is the winner-neighbor advantage in those iterations minus the winner-neighbor advantage in even iterations 2 to 12, when the defender has not changed since the previous iteration. In the frozen arm, the same comparison shows the difference between odd and even iterations without any defender updates.
    \item \emph{Common opening} (M2, M3): the four-word opening shared by the largest number of valid training attacks of an iteration. The common-opening advantage is the difference in success rates between attacks that carry the opening and other attacks. Unlike training advantage, the winner-neighbor and common-opening advantages compare success rates. The share of that opening and the mean training advantage of each group also summarize the dominant pattern of the attacker.
    \item \emph{Row-cap diagnostics} (M4): the enrichment of the common opening among selected rows relative to all valid attacks, the diversity of selected rows relative to all valid attacks of the same behaviors, and, for the retained runs, the difference between the family mix of selected rows and of all signal rows.
    \item \emph{Update size} (M5): the mean KL to the base model and the largest gradient norm of each attacker update.
    \item \emph{Length and wording} (M6): mean attack length in characters, length-adjusted ADR (which averages pair distances within length bands represented in every bank and combines the bands with fixed weights), and mean pairwise Jaccard distance between word 3-gram sets of the first 120 words.
\end{itemize}

\subsection{Pre-Specification, Exploratory Status and Containment}

The arms, the shared seed, the settings and the two primary readings were written down in version control before the first iteration. The continuation from iteration 12 to iteration 24 started after the readings at iteration 12 were known, so every reading beyond iteration 12 is exploratory. The mechanism record, the equal-progress comparisons, the length-adjusted readings and the analyses of the retained runs are exploratory as well. All readings are uncorrected for multiple comparisons, and uncertainty intervals reflect variation across behaviors only. Success is the classifier decision. A human audit and an independent verifier have not been collected, and held-out test splits remain unscored.

This study excludes production targets, functional payloads and external model APIs. Raw benchmark text, attacks, answers, preference examples and adapters remain private, and the analyses use numerical summaries.

\section{Results}
\label{sec:results}

\subsection{Trajectories of Diversity and Success}

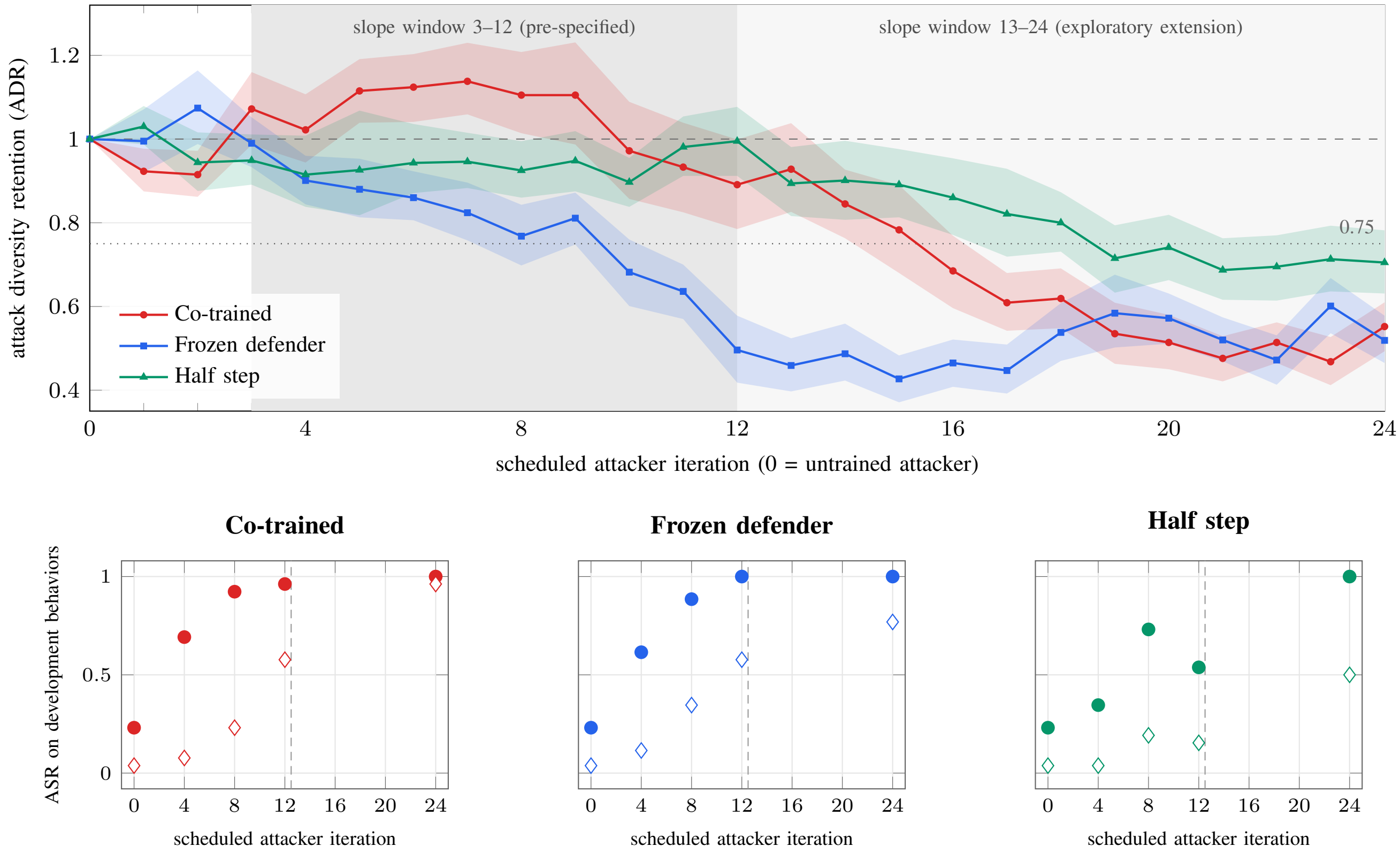
\begin{figure*}[t]
\centering
\definecolor{coC}{HTML}{DC2626}\definecolor{fdC}{HTML}{2563EB}\definecolor{hsC}{HTML}{059669}
\begin{tikzpicture}
\begin{axis}[
    width=0.97\textwidth, height=6.6cm,
    xmin=0, xmax=24, ymin=0.35, ymax=1.32,
    xtick={0,4,8,12,16,20,24}, minor x tick num=3,
    ytick={0.4,0.6,0.8,1.0,1.2},
    yticklabel style={/pgf/number format/fixed,/pgf/number format/precision=1},
    xlabel={scheduled attacker iteration (0 = untrained attacker)}, ylabel={attack diversity retention (ADR)},
    grid=major, grid style={black!8},
    tick label style={font=\footnotesize}, label style={font=\footnotesize},
    legend style={at={(0.015,0.03)},anchor=south west,font=\footnotesize,draw=none,fill=white,fill opacity=0.85,text opacity=1,legend cell align=left},
    clip=true]
  \addplot[name path=w1Top,draw=none,forget plot] coordinates {(3,1.32) (12,1.32)};
  \addplot[name path=w1Bot,draw=none,forget plot] coordinates {(3,0.35) (12,0.35)};
  \addplot[fill=black!9,draw=none,forget plot] fill between[of=w1Top and w1Bot];
  \addplot[name path=w2Top,draw=none,forget plot] coordinates {(12,1.32) (24,1.32)};
  \addplot[name path=w2Bot,draw=none,forget plot] coordinates {(12,0.35) (24,0.35)};
  \addplot[fill=black!3,draw=none,forget plot] fill between[of=w2Top and w2Bot];
  \node[font=\scriptsize,anchor=north,text=black!70] at (axis cs:7.5,1.31) {slope window 3--12 (pre-specified)};
  \node[font=\scriptsize,anchor=north,text=black!70] at (axis cs:18,1.31) {slope window 13--24 (exploratory extension)};
  \draw[dashed,black!55] (axis cs:0,1) -- (axis cs:24,1);
  \draw[dotted,black!70] (axis cs:0,0.75) -- (axis cs:24,0.75);
  \node[font=\scriptsize,anchor=south east,text=black!60] at (axis cs:24,0.75) {0.75};
  \addplot[name path=coHi,draw=none,forget plot] coordinates {(0,1.000) (1,0.977) (2,0.972) (3,1.160) (4,1.107) (5,1.191) (6,1.203) (7,1.230) (8,1.208) (9,1.231) (10,1.089) (11,1.039) (12,0.999) (13,1.038) (14,0.927) (15,0.890) (16,0.769) (17,0.680) (18,0.691) (19,0.609) (20,0.580) (21,0.529) (22,0.562) (23,0.527) (24,0.610)};
  \addplot[name path=coLo,draw=none,forget plot] coordinates {(0,1.000) (1,0.875) (2,0.862) (3,0.986) (4,0.944) (5,1.039) (6,1.041) (7,1.059) (8,1.014) (9,0.987) (10,0.857) (11,0.825) (12,0.785) (13,0.826) (14,0.763) (15,0.680) (16,0.596) (17,0.542) (18,0.548) (19,0.463) (20,0.450) (21,0.421) (22,0.465) (23,0.412) (24,0.493)};
  \addplot[coC,fill opacity=0.16,draw=none,forget plot] fill between[of=coHi and coLo];
  \addplot[coC,thick,mark=*,mark size=0.9pt] coordinates {(0,1.000) (1,0.923) (2,0.915) (3,1.072) (4,1.022) (5,1.115) (6,1.124) (7,1.138) (8,1.105) (9,1.105) (10,0.972) (11,0.933) (12,0.891) (13,0.928) (14,0.845) (15,0.783) (16,0.685) (17,0.609) (18,0.619) (19,0.535) (20,0.514) (21,0.476) (22,0.514) (23,0.468) (24,0.552)};
  \addlegendentry{Co-trained}
  \addplot[name path=fdHi,draw=none,forget plot] coordinates {(0,1.000) (1,1.071) (2,1.164) (3,1.052) (4,0.960) (5,0.953) (6,0.923) (7,0.896) (8,0.843) (9,0.872) (10,0.760) (11,0.700) (12,0.578) (13,0.524) (14,0.559) (15,0.483) (16,0.521) (17,0.509) (18,0.608) (19,0.676) (20,0.631) (21,0.574) (22,0.531) (23,0.668) (24,0.578)};
  \addplot[name path=fdLo,draw=none,forget plot] coordinates {(0,1.000) (1,0.921) (2,0.988) (3,0.933) (4,0.844) (5,0.813) (6,0.806) (7,0.758) (8,0.698) (9,0.748) (10,0.601) (11,0.570) (12,0.418) (13,0.397) (14,0.423) (15,0.371) (16,0.408) (17,0.392) (18,0.470) (19,0.502) (20,0.511) (21,0.469) (22,0.413) (23,0.536) (24,0.465)};
  \addplot[fdC,fill opacity=0.16,draw=none,forget plot] fill between[of=fdHi and fdLo];
  \addplot[fdC,thick,mark=square*,mark size=0.8pt] coordinates {(0,1.000) (1,0.995) (2,1.074) (3,0.990) (4,0.901) (5,0.880) (6,0.860) (7,0.824) (8,0.768) (9,0.811) (10,0.682) (11,0.636) (12,0.496) (13,0.459) (14,0.487) (15,0.427) (16,0.465) (17,0.447) (18,0.538) (19,0.584) (20,0.572) (21,0.520) (22,0.472) (23,0.601) (24,0.519)};
  \addlegendentry{Frozen defender}
  \addplot[name path=hsHi,draw=none,forget plot] coordinates {(0,1.000) (1,1.079) (2,1.016) (3,1.011) (4,1.008) (5,1.068) (6,1.036) (7,1.015) (8,0.995) (9,1.019) (10,0.955) (11,1.054) (12,1.077) (13,0.981) (14,0.996) (15,0.976) (16,0.954) (17,0.929) (18,0.873) (19,0.794) (20,0.819) (21,0.763) (22,0.770) (23,0.793) (24,0.782)};
  \addplot[name path=hsLo,draw=none,forget plot] coordinates {(0,1.000) (1,0.987) (2,0.876) (3,0.891) (4,0.838) (5,0.818) (6,0.871) (7,0.883) (8,0.860) (9,0.875) (10,0.838) (11,0.912) (12,0.912) (13,0.816) (14,0.807) (15,0.813) (16,0.771) (17,0.719) (18,0.731) (19,0.633) (20,0.663) (21,0.616) (22,0.614) (23,0.636) (24,0.631)};
  \addplot[hsC,fill opacity=0.16,draw=none,forget plot] fill between[of=hsHi and hsLo];
  \addplot[hsC,thick,mark=triangle*,mark size=1.0pt] coordinates {(0,1.000) (1,1.030) (2,0.944) (3,0.949) (4,0.915) (5,0.926) (6,0.943) (7,0.946) (8,0.925) (9,0.948) (10,0.897) (11,0.981) (12,0.995) (13,0.894) (14,0.901) (15,0.891) (16,0.860) (17,0.821) (18,0.800) (19,0.715) (20,0.741) (21,0.687) (22,0.695) (23,0.713) (24,0.705)};
  \addlegendentry{Half step}
\end{axis}
\end{tikzpicture}

\medskip
\definecolor{successCo}{HTML}{DC2626}
\definecolor{successFrozen}{HTML}{2563EB}
\definecolor{successHalf}{HTML}{059669}
\begin{tikzpicture}
\pgfplotsset{
  success checkpoint plot/.style={
    width=0.31\textwidth, height=4.4cm,
    xmin=-1, xmax=25, ymin=-0.08, ymax=1.08,
    xtick={0,4,8,12,16,20,24}, ytick={0,0.5,1},
    xlabel={scheduled attacker iteration},
    tick label style={font=\scriptsize}, label style={font=\scriptsize},
    title style={font=\small\bfseries}, grid=major, grid style={black!10},
    axis line style={black!60}, tick style={black!60}
  }
}
\begin{axis}[success checkpoint plot,title={Co-trained},ylabel={ASR on development behaviors}]
  \draw[densely dashed,black!40] (axis cs:12.5,-0.08) -- (axis cs:12.5,1.08);
  \addplot[only marks,successCo,mark=*,mark size=2.2pt] coordinates {(0,0.231) (4,0.692) (8,0.923) (12,0.962) (24,1.000)};
  \addplot[only marks,successCo,mark=diamond*,mark size=2.7pt,mark options={fill=white}] coordinates {(0,0.038) (4,0.077) (8,0.231) (12,0.577) (24,0.962)};
\end{axis}
\begin{axis}[success checkpoint plot,title={Frozen defender},at={(5.65cm,0)},yticklabels={,,}]
  \draw[densely dashed,black!40] (axis cs:12.5,-0.08) -- (axis cs:12.5,1.08);
  \addplot[only marks,successFrozen,mark=*,mark size=2.2pt] coordinates {(0,0.231) (4,0.615) (8,0.885) (12,1.000) (24,1.000)};
  \addplot[only marks,successFrozen,mark=diamond*,mark size=2.7pt,mark options={fill=white}] coordinates {(0,0.038) (4,0.115) (8,0.346) (12,0.577) (24,0.769)};
\end{axis}
\begin{axis}[success checkpoint plot,title={Half step},at={(11.3cm,0)},yticklabels={,,}]
  \draw[densely dashed,black!40] (axis cs:12.5,-0.08) -- (axis cs:12.5,1.08);
  \addplot[only marks,successHalf,mark=*,mark size=2.2pt] coordinates {(0,0.231) (4,0.346) (8,0.731) (12,0.538) (24,1.000)};
  \addplot[only marks,successHalf,mark=diamond*,mark size=2.7pt,mark options={fill=white}] coordinates {(0,0.038) (4,0.038) (8,0.192) (12,0.154) (24,0.500)};
\end{axis}
\end{tikzpicture}
\caption{Diversity and success by scheduled attacker iteration (0 is the untrained attacker). Top: ADR point estimates and pointwise 95\% paired behavior-bootstrap intervals. The dashed line marks ADR 1, and the dotted line marks the 0.75 ADR reference. Shading marks the slope windows: iterations 3--12 (pre-specified) and 13--24 (exploratory). Bottom: ASR@8 (filled circles) and ASR@1 (open diamonds) against the untrained defender in every arm. Only the documented success checkpoints 0, 4, 8, 12 and 24 are shown. Points are not connected, and intervals are not shown. The vertical dashed line separates the exploratory continuation. Each arm has one seed. The development panel has 26 behaviors.}
\label{fig:adr}
\end{figure*}

All three arms completed 24 iterations. Figure~\ref{fig:adr} shows ADR and selected success checkpoints. Table~\ref{tab:primary} reports the primary ADR readings.

\begin{table*}[t]
\centering
\caption{Cumulative change and slope of ADR, with 95\% behavior-bootstrap intervals in parentheses. Contrasts are differences from the co-trained arm. Iteration-12 readings were pre-specified. Iteration-24 readings are exploratory. One seed per arm.}
\label{tab:primary}
\small
\begin{tabular}{@{}P{0.32\textwidth}P{0.20\textwidth}P{0.20\textwidth}P{0.20\textwidth}@{}}
\toprule
Reading & Co-trained & Frozen defender & Half step \\
\midrule
\multicolumn{4}{@{}l}{\emph{Pre-specified horizon: 12 iterations}} \\
ADR at iteration 12 & 0.891 & 0.496 & 0.995 \\
Change from untrained attacker & $-0.109$ ($-0.209$, $0.002$) & $-0.504$ ($-0.581$, $-0.417$) & $-0.005$ ($-0.086$, $0.077$) \\
Slope per iteration, iterations 3--12 & $-0.019$ ($-0.030$, $-0.007$) & $-0.045$ ($-0.057$, $-0.034$) & $+0.005$ ($-0.006$, $0.016$) \\
Contrast, change & --- & $-0.395$ ($-0.520$, $-0.270$) & $+0.104$ ($-0.0003$, $0.206$) \\
Contrast, slope & --- & $-0.027$ ($-0.039$, $-0.015$) & $+0.022$ ($0.007$, $0.037$) \\
\addlinespace
\multicolumn{4}{@{}l}{\emph{Exploratory extension: 24 iterations}} \\
ADR at iteration 24 & 0.552 (0.493, 0.610) & 0.519 (0.465, 0.578) & 0.705 (0.631, 0.782) \\
Change from untrained attacker & $-0.448$ ($-0.508$, $-0.391$) & $-0.481$ ($-0.534$, $-0.421$) & $-0.295$ ($-0.370$, $-0.218$) \\
Slope per iteration, iterations 13--24 & $-0.038$ ($-0.046$, $-0.031$) & $+0.009$ ($0.005$, $0.014$) & $-0.022$ ($-0.030$, $-0.014$) \\
Contrast, change & --- & $-0.033$ ($-0.088$, $0.020$) & $+0.153$ ($0.068$, $0.228$) \\
Contrast, slope & --- & $+0.047$ ($0.039$, $0.056$) & $+0.016$ ($0.007$, $0.025$) \\
\addlinespace
\multicolumn{4}{@{}l}{\emph{Timing and success}} \\
First iteration with ADR below 0.75 & 16 & 10 & 19 \\
First iteration with the whole interval below 0.75 & 17 & 11 & $>24$ \\
ADR on a fresh bank at iteration 24 & 0.516 & 0.562 & 0.714 \\
Success fraction, iterations 12 and 24 & 0.500, 0.837 & 0.611, 0.673 & 0.163, 0.534 \\
\bottomrule
\end{tabular}
\end{table*}

The frozen-defender arm narrows first. At iteration 12, its ADR is 0.496, a decrease of 0.504 from the untrained ADR of 1. The co-trained arm has an ADR of 0.891, a decrease of 0.109, and the half-step arm has an ADR of 0.995, a decrease of 0.005. The contrast of the frozen arm with the co-trained arm is $-0.395$ (95\% interval $-0.520$ to $-0.270$) in cumulative change and $-0.027$ per iteration in slope. The point estimate of ADR first decreases below 0.75 at iteration 10 (frozen), 16 (co-trained) and 19 (half step). The whole interval lies below 0.75 from iteration 11, 17 and beyond iteration 24, respectively. The co-trained arm first rises: ADR exceeds 1 over iterations 3 to 9 and peaks at 1.138 at iteration 7 before the fall.

The co-trained and frozen arms end close together. At iteration 24 the cumulative change is $-0.448$ (co-trained) and $-0.481$ (frozen), a contrast of $-0.033$ ($-0.088$ to $0.020$), and mean ADR over iterations 21 to 24 is 0.503 and 0.528. The frozen arm reaches 0.427 at iteration 15, rises to between 0.47 and 0.60 over iterations 18 to 24 and levels off, so the iteration-12 value was a point on the way. The half-step arm ends at 0.705 (0.631 to 0.782), 0.153 above the co-trained arm (0.068 to 0.228), and stays between 0.69 and 0.71 over iterations 21 to 24. The newly generated banks at iteration 24 also put the co-trained and frozen arms below the half-step arm, with ADR 0.516 for co-training, 0.562 for the frozen defender and 0.714 for half step (Table~\ref{tab:primary}). The whole interval of the co-trained arm lies below 0.75 on the fresh banks at iterations 16, 20 and 24 (0.636, 0.466 and 0.516). The corresponding upper interval ends are 0.725, 0.527 and 0.568, whereas the half-step intervals reach 0.926, 0.777 and 0.790.

By iteration 24, success is higher and ADR lower in every arm, but success can rise before diversity falls. At iteration 8 the co-trained arm reaches ASR@8 of 0.923 while ADR is 1.105 (95\% interval 1.014 to 1.208). At iteration 12 ASR@8 is 0.962 and 1.000 in the co-trained and frozen arms, while ADR is 0.891 and 0.496. Their success fractions differ (0.500 and 0.611), so the two readings do not compare the arms at the same level of attack success. At iteration 24 the success fraction is 0.837, 0.673 and 0.534, ASR@1 is 0.962, 0.769 and 0.500, and ASR@8 is 1.000 in every arm (untrained: 0.043, 0.038 and 0.231). Each ASR@1 success represents one of 26 behaviors, so its resolution is about 0.038.
At iteration 12, the success fraction is 0.500 for the co-trained arm, 0.611 for the frozen-defender arm and 0.163 for the half-step arm. To compare diversity at similar levels of success rather than at the same iteration, an exploratory analysis groups checkpoints by success fraction and averages their ADR values within each group. For success fractions from 0.4 to 0.5, mean ADR is 0.93 for the co-trained arm, 0.68 for the frozen-defender arm and 0.75 for the half-step arm. For success fractions from 0.5 to 0.6, mean ADR is 0.87 for the co-trained arm, 0.64 for the frozen-defender arm and 0.71 for the half-step arm. Each arm has only one to four checkpoints in each range. At success fractions of 0.7 and above, mean ADR is 0.55 across nine co-trained checkpoints and 0.51 across six frozen-defender checkpoints. The co-trained and frozen-defender arms therefore have similar mean ADR at high success, while the half-step arm lies between them at intermediate success. The half-step arm retains nearly all its initial diversity at iteration 12, when its attacks are less successful than those in the other arms. Its mean ADR is lower in the 0.4 to 0.6 success-fraction ranges.

\subsection{Winners Are Narrower than All Valid Attacks}

\begin{figure*}[t]
\centering
\begin{tikzpicture}[x=1cm,y=1cm]
  \definecolor{poolBlue}{HTML}{94A3B8}
  \definecolor{winnerRed}{HTML}{DC2626}
  \definecolor{poolLine}{HTML}{CBD5E1}
  \tikzset{
    pool/.style={circle,fill=poolBlue,inner sep=2.1pt},
    win/.style={circle,fill=winnerRed,inner sep=2.1pt},
    candidate link/.style={draw=poolLine,line width=0.35pt},
    winner link/.style={draw=winnerRed!35,line width=0.4pt}
  }
  \newcommand{\candidatecloud}[2]{%
    \begin{scope}[shift={(#1,#2)}]
      \coordinate (p1) at (-1,0.2);
      \coordinate (p2) at (0,0.95);
      \coordinate (p3) at (1,0.2);
      \coordinate (p4) at (0,-0.95);
      \coordinate (p5) at (-0.65,0.55);
      \coordinate (p6) at (0.65,0.55);
      \coordinate (p7) at (0.65,-0.55);
      \coordinate (p8) at (-0.65,-0.55);
      \foreach \i in {1,...,8} {
        \foreach \j in {1,...,8} {
          \ifnum\i<\j
            \draw[candidate link] (p\i) -- (p\j);
          \fi
        }
      }
      \foreach \i in {1,...,4} { \node[pool] at (p\i) {}; }
      \foreach \i in {5,...,8} { \node[win] at (p\i) {}; }
    \end{scope}
  }
  \newcommand{\winnercloud}[2]{%
    \begin{scope}[shift={(#1,#2)}]
      \coordinate (w1) at (-0.65,0.55);
      \coordinate (w2) at (0.65,0.55);
      \coordinate (w3) at (0.65,-0.55);
      \coordinate (w4) at (-0.65,-0.55);
      \foreach \i in {1,...,4} {
        \foreach \j in {1,...,4} {
          \ifnum\i<\j
            \draw[winner link] (w\i) -- (w\j);
          \fi
        }
      }
      \foreach \i in {1,...,4} { \node[win] at (w\i) {}; }
    \end{scope}
  }

  \node[font=\small\bfseries,align=center] at (-3,1.65) {All valid\\attacks};
  \node[font=\small\bfseries,align=center] at (3,1.65) {All successful\\attacks};
  \node[font=\small\bfseries,anchor=east] at (-4.65,0) {Update 1};
  \candidatecloud{-3}{0}
  \winnercloud{3}{0}
  \node[font=\scriptsize] at (-3,-1.25) {mean distance 0.217};
  \node[font=\scriptsize] at (3,-1.25) {mean distance 0.181};

  \node[font=\small\bfseries,anchor=east] at (-4.65,-3.1) {Update 24};
  \candidatecloud{-3}{-3.1}
  \winnercloud{3}{-3.1}
  \node[font=\scriptsize] at (-3,-4.35) {mean distance 0.147};
  \node[font=\scriptsize] at (3,-4.35) {mean distance 0.127};

  \node[pool,anchor=west] at (-4,-5.05) {};
  \node[font=\scriptsize,align=left,anchor=west] at (-3.8,-5.05) {other valid\\attacks};
  \node[win,anchor=west] at (2,-5.05) {};
  \node[font=\scriptsize,align=left,anchor=west] at (2.2,-5.05) {successful\\attacks};
\end{tikzpicture}
\caption{Winner dispersion illustrated with measured values from retained run 1011. Each row represents one update: the left cloud shows valid attacks with successful attacks highlighted, and the right cloud shows those successful attacks alone. Distances are mean pairwise cosine distances averaged within behavior over the same behaviors with at least two winners: 15 behaviors at update 1 and 89 at update 24. The eligible behaviors differ between updates. The ratio is winner distance divided by attack distance. Dot positions are schematic, not embedding coordinates. This illustrates winner dispersion, not ADR or a before-and-after causal effect. Likelihood changes of fixed attacks were not measured.}
\label{fig:winner-dispersion}
\end{figure*}
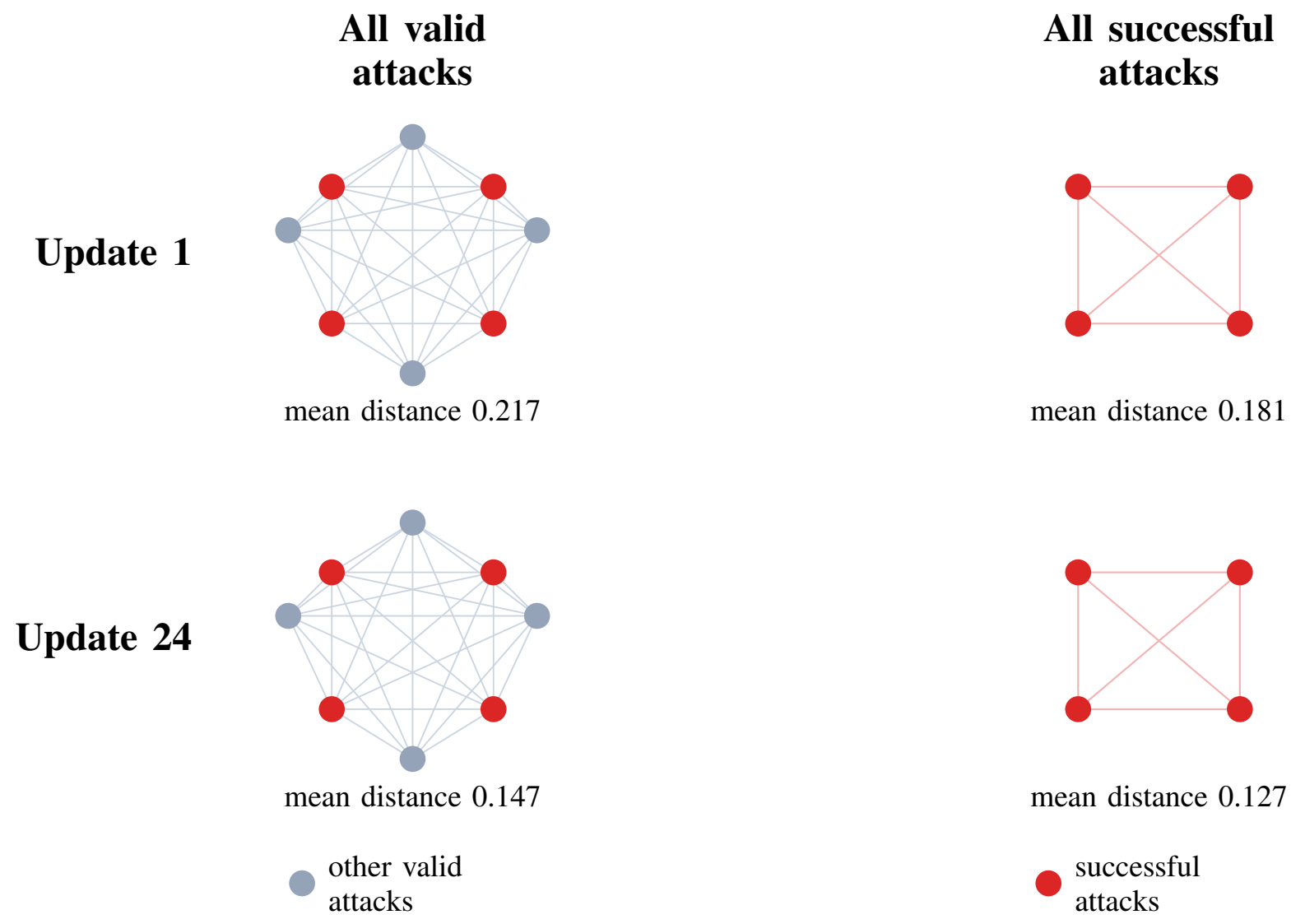

M3 implies that attacks receiving positive training advantage should be narrower than the distribution of all generated attacks. Winner dispersion provides a related descriptive reading of all successful attacks, including winners with zero training advantage. Figure~\ref{fig:winner-dispersion} illustrates the comparison at two updates from one retained run. In the retained runs, the mean pairwise distance among winners lies below that among all valid attacks in 72 of 72 update-level comparisons (behaviors with at least two winners, 5 to 92 of 96 behaviors per update). The ratio of the two is 0.730, 0.773 and 0.873 over updates 1 to 3 and 0.800, 0.873 and 0.915 over updates 22 to 24 in the three runs. In the arms the ratio ranges over 0.70 to 0.92 across iterations 1 to 12 with a flat course (Table~\ref{tab:records}) and stays between 0.70 and 0.93 across all 24 iterations. Winners are thus narrower than the pool at the first iteration, and the gap persists while the pool itself narrows. The 72 comparisons are correlated observations from three evolving policies and fall short of 72 independent confirmations. In the last update of all three runs the winners also lie below the pool within the longest length band, a coarse check on thin length support.

\begin{table*}[t]
\centering
\caption{Mechanism records by arm. Parentheses give 95\% behavior-bootstrap intervals. Ranges give the minimum and maximum over the iterations listed. Entries separated by a slash give the common opening and the other attacks. \textsuperscript{$\dagger$}The frozen arm has a fixed defender, so the entry shows the contrast produced by iteration parity alone.}
\label{tab:records}
\small
\begin{tabular}{@{}P{0.32\textwidth}P{0.20\textwidth}P{0.20\textwidth}P{0.20\textwidth}@{}}
\toprule
Record & Co-trained & Frozen defender & Half step \\
\midrule
(M3) Winner dispersion ratio, iterations 1--12 & 0.70--0.82 & 0.79--0.92 & 0.72--0.87 \\
(M2) Winner-neighbor advantage, iterations 2--12 & $+0.133$ (0.106, 0.159) & $+0.135$ (0.108, 0.162) & $+0.138$ (0.111, 0.163) \\
(M2) Relative risk, iterations 2--12 & 1.64 & 1.55 & 1.80 \\
(M2) Winner-neighbor advantage, iterations 13--24 & $+0.104$ & $+0.110$ & $+0.143$ \\
(M1) Winner-neighbor advantage after a defender update minus that between defender updates, iterations 2--12 & $+0.014$ ($-0.026$, $0.053$) & $-0.012$ ($-0.052$, $0.034$)\textsuperscript{$\dagger$} & $-0.012$ ($-0.052$, $0.025$) \\
(M2) Common-opening share of training attacks, iterations 21--24 & 0.83--0.88 & 0.80--0.92 & 0.62--0.77 \\
(M2) Success share of common opening / others, iterations 7--12 & 0.35 / 0.21 & 0.31 / 0.26 & 0.20 / 0.13 \\
(M2) Success share of common opening / others, iterations 19--24 & 0.62 / 0.52 & 0.60 / 0.47 & 0.33 / 0.23 \\
(M1) Success share of training attacks, iterations 21--24 & 0.57--0.65 & 0.51--0.67 & 0.25--0.35 \\
(M5) Largest mean KL to the base model through iteration 19 (iteration) & 979 (18) & 2,562 (2) & 1.2 \\
(M6) Mean attack length at iteration 24, characters (untrained 221) & 311 & 418 & 367 \\
(M6) ADR at iteration 24, after length adjustment & 0.68 (0.54, 0.83) & 0.57 (0.50, 0.66) & 0.84 (0.70, 0.96) \\
(M6) ADR at iteration 24, unadjusted & 0.552 & 0.519 & 0.705 \\
(M6) Within-behavior distance at iteration 24 (untrained 0.211) & 0.109 & 0.108 & 0.151 \\
\bottomrule
\end{tabular}
\end{table*}

Family labels show one attack family becoming dominant. In the retained runs, hypothetical framing accounts for 6.9\%, 6.9\% and 8.2\% of the available attacks over updates 1 to 3 and for 74.3\%, 62.0\% and 41.1\% over updates 22 to 24, and the share among winners starts higher (18.4\%, 14.5\% and 10.5\%).

The observations of narrowing are not fully explained by either of the following two aspects. First, narrowing does not always accompany a shift toward a more concentrated mix of strategy families. The Herfindahl--Hirschman index measures concentration as the sum of squared family shares, with higher values indicating a less even mix. In one retained run, the index falls from 0.400 at the first update to 0.277 at the last, while the mean embedding distance between attacks of the same family and behavior falls from 0.208 to 0.130. Attacks within families become closer in embedding space even as the family mix becomes more balanced. Second, narrowing continues after groups of 16 attacks with only one winner contribute little of the positive training advantage. Across the three retained runs, the share contributed by these groups falls from 46.7\%, 66.4\% and 45.2\% over the first three updates to 2.4\%, 7.6\% and 6.0\% over the last three. Other factors may also contribute to the continued decline in diversity. The influence of single-winner groups earlier in the run may still affect diversity later.

\subsection{Success of Attacks Similar to Earlier Winners}

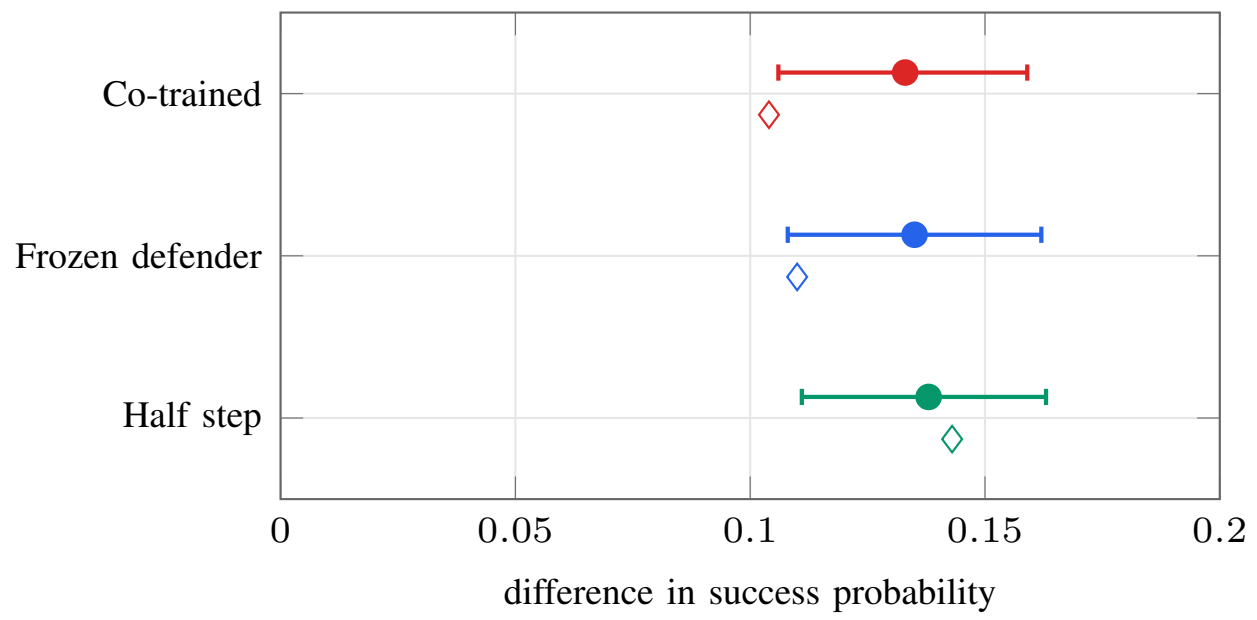
\begin{figure}[t]
\centering
\definecolor{simCo}{HTML}{DC2626}
\definecolor{simFrozen}{HTML}{2563EB}
\definecolor{simHalf}{HTML}{059669}
\begin{tikzpicture}
\begin{axis}[
    width=0.88\columnwidth, height=4.8cm,
    xmin=0, xmax=0.20, ymin=0.5, ymax=3.5,
    xtick={0,0.05,0.10,0.15,0.20},
    xticklabel style={/pgf/number format/fixed,/pgf/number format/precision=2},
    ytick={1,2,3}, yticklabels={Half step,Frozen defender,Co-trained},
    xlabel={difference in success probability},
    tick label style={font=\scriptsize}, label style={font=\scriptsize},
    grid=major, grid style={black!10}, axis line style={black!60},
    tick style={black!60}, clip=false]
  \draw[simCo,thick] (axis cs:0.106,3.13) -- (axis cs:0.159,3.13);
  \draw[simCo,thick] (axis cs:0.106,3.08) -- (axis cs:0.106,3.18);
  \draw[simCo,thick] (axis cs:0.159,3.08) -- (axis cs:0.159,3.18);
  \draw[simFrozen,thick] (axis cs:0.108,2.13) -- (axis cs:0.162,2.13);
  \draw[simFrozen,thick] (axis cs:0.108,2.08) -- (axis cs:0.108,2.18);
  \draw[simFrozen,thick] (axis cs:0.162,2.08) -- (axis cs:0.162,2.18);
  \draw[simHalf,thick] (axis cs:0.111,1.13) -- (axis cs:0.163,1.13);
  \draw[simHalf,thick] (axis cs:0.111,1.08) -- (axis cs:0.111,1.18);
  \draw[simHalf,thick] (axis cs:0.163,1.08) -- (axis cs:0.163,1.18);
  \addplot[only marks,mark=*,mark size=2.3pt,simCo] coordinates {(0.133,3.13)};
  \addplot[only marks,mark=*,mark size=2.3pt,simFrozen] coordinates {(0.135,2.13)};
  \addplot[only marks,mark=*,mark size=2.3pt,simHalf] coordinates {(0.138,1.13)};
  \addplot[only marks,mark=diamond*,mark size=2.5pt,simCo,mark options={fill=white}] coordinates {(0.104,2.87)};
  \addplot[only marks,mark=diamond*,mark size=2.5pt,simFrozen,mark options={fill=white}] coordinates {(0.110,1.87)};
  \addplot[only marks,mark=diamond*,mark size=2.5pt,simHalf,mark options={fill=white}] coordinates {(0.143,0.87)};
\end{axis}
\end{tikzpicture}
\caption{Winner-neighbor advantage: success-rate difference for attacks whose nearest neighbor in the previous iteration was a winner compared with other attacks. Filled circles show iterations 2--12, with 95\% behavior-bootstrap intervals. Open diamonds show iterations 13--24, for which intervals are not shown. Positive values mean higher success probability than for attacks whose nearest neighbor was not a winner. Each arm has one seed. Both windows are exploratory, and the association does not identify a cause.}
\label{fig:winner-similarity}
\end{figure}

An attack whose nearest neighbor for the same behavior at the previous iteration was a winner succeeds more often than an attack whose nearest neighbor was not a winner (Figure~\ref{fig:winner-similarity}). Combining the success-rate differences across behaviors and iterations 2 to 12 gives winner-neighbor advantages of $+0.133$, $+0.135$ and $+0.138$ in the co-trained, frozen and half-step arms, with $n=16{,}840$, $16{,}840$ and $16{,}374$ attacks (Table~\ref{tab:records}). Across iterations 13 to 24, the corresponding winner-neighbor advantages are $+0.104$, $+0.110$ and $+0.143$. If successful and unsuccessful earlier attacks were equally likely to be the nearest neighbor for the same behavior, the share of new attacks with a winner as nearest neighbor would reflect the earlier success share. In the co-trained arm over iterations 2 to 6, 13.4\% of attacks have a previous winner as their nearest neighbor, compared with 9.3\% expected from the earlier success share. Across arms and the two windows of iterations 2 to 12, the observed share is 1.28 to 1.57 times the expected share.

\begingroup
\setlength{\parskip}{0pt}
Under M1, the winner-neighbor advantage should be smaller after a defender update, because a changed defender changes which attacks succeed. The contrast between iterations that follow a defender update and iterations within a fixed defender is $+0.014$ ($-0.026$ to $0.053$) in the co-trained arm ($+0.141$ against $+0.127$) and $-0.012$ ($-0.052$ to $0.025$) at half step. The frozen arm, with a fixed defender, gives $-0.012$ ($-0.052$ to $0.034$), the size produced by iteration parity alone. Over iterations 13 to 24 the winner-neighbor advantage is $+0.112$ after a defender update against $+0.120$ within a fixed defender in the co-trained arm, $+0.132$ against $+0.149$ at half step, and $+0.118$ against $+0.122$ in the frozen arm. The intervals rule out a fall of the winner-neighbor advantage across a defender update larger than 0.03 (co-trained) or 0.05 (half step) and leave smaller changes open. The winner-neighbor advantage of 0.13 to 0.14 over iterations 2 to 12 is the same in every arm, including the arm with a fixed defender.\par
\endgroup

\subsection{A Dominant Attack Pattern Takes Over}

\begin{figure*}[t]
\centering
\definecolor{openingCommon}{HTML}{DC2626}
\definecolor{openingOther}{HTML}{475569}
\begin{tikzpicture}
\pgfplotsset{
  opening plot/.style={
    width=0.31\textwidth, height=5.0cm,
    xmin=0.75, xmax=3.25, ymin=0, ymax=0.70,
    xtick={1,2,3}, xticklabels={7--12,13--18,19--24},
    ytick={0,0.2,0.4,0.6},
    xlabel={attacker iterations},
    tick label style={font=\scriptsize}, label style={font=\scriptsize},
    title style={font=\small\bfseries}, grid=major, grid style={black!10},
    axis line style={black!60}, tick style={black!60}
  }
}
\begin{axis}[opening plot,title={Co-trained},ylabel={mean success fraction}]
  \addplot[openingOther,thick,dashed,mark=o,mark size=2pt,mark options={fill=white}] coordinates {(1,0.21) (2,0.36) (3,0.52)};
  \addplot[openingCommon,thick,mark=*,mark size=2pt] coordinates {(1,0.35) (2,0.52) (3,0.62)};
\end{axis}
\begin{axis}[opening plot,title={Frozen defender},at={(5.65cm,0)},yticklabels={,, ,}]
  \addplot[openingOther,thick,dashed,mark=o,mark size=2pt,mark options={fill=white}] coordinates {(1,0.26) (2,0.42) (3,0.47)};
  \addplot[openingCommon,thick,mark=*,mark size=2pt] coordinates {(1,0.31) (2,0.52) (3,0.60)};
\end{axis}
\begin{axis}[opening plot,title={Half step},at={(11.3cm,0)},yticklabels={,, ,}]
  \addplot[openingOther,thick,dashed,mark=o,mark size=2pt,mark options={fill=white}] coordinates {(1,0.13) (2,0.17) (3,0.23)};
  \addplot[openingCommon,thick,mark=*,mark size=2pt] coordinates {(1,0.20) (2,0.27) (3,0.33)};
\end{axis}
\end{tikzpicture}
\caption{Success of training attacks with the most common four-word opening (red, solid circles) and other openings (gray, open circles). Each point is the mean success fraction over the six iterations in its window. The most common opening is determined separately at each iteration and may change. The defender is the current training defender of each arm. These exploratory descriptive comparisons have one seed per arm, no intervals shown, and do not establish that the opening causes success.}
\label{fig:opening-success}
\end{figure*}
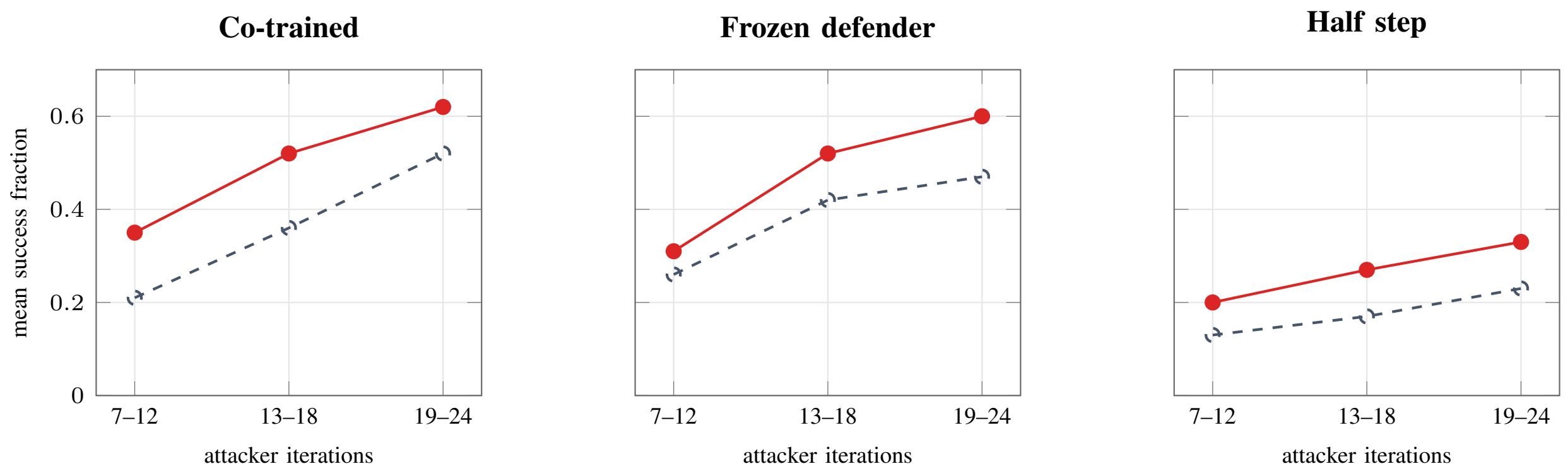

Within each iteration, the most common four-word opening among valid training attacks defines the dominant pattern of the attacker. This attack opening accounts for 0.83 to 0.88 of valid training attacks over iterations 21 to 24 in the co-trained arm (0.07 to 0.16 over iterations 7 to 11), 0.80 to 0.92 in the frozen arm (0.26 at iteration 1 and 0.66 at iteration 12) and 0.62 to 0.77 at half step. The common-opening advantage is 0.05 to 0.16 in every arm iteration 7 (Figure~\ref{fig:opening-success} and Table~\ref{tab:records}), while the alternatives still succeed in 0.47 to 0.52 of the cases late in the co-trained and frozen arms.

In the co-trained arm, training advantages first favor attacks with the common opening and later disfavor alternatives. Attacks with the common opening carry a mean training advantage of $+0.07$ to $+0.28$ over iterations 9 to 14, against $-0.01$ to $-0.07$ for attacks without that opening. From iteration 17 the mean training advantage of attacks with the common opening is near zero, and attacks without it carry $-0.12$ to $-0.18$ over iterations 17 to 21. The frozen arm shows the attacks without the opening carry $-0.09$ to $-0.16$ at iterations 16 and 20 to 23.

The mean cosine distance between embeddings of attacks for the same behavior falls from 0.211 for the untrained attacker to 0.109, 0.108 and 0.151 at iteration 24 in the co-trained, frozen and half-step arms, respectively. The mean distance between attacks for different behaviors also falls, from 0.232 to 0.136, 0.124 and 0.178. The between-behavior distance exceeds the within-behavior distance by 15\% to 25\% at iteration 24 against 10\% untrained, so the attacker writes nearly the same attack again for a behavior, and the dominant attack pattern is shared across behaviors. The dominant family differs between arms: hypothetical framing reaches 0.57 to 0.62 of the development bank at iterations 21 to 24 in the co-trained arm and 0.32 to 0.38 at half step (0.05 untrained), whereas the frozen arm alternates between hypothetical framing and transformation.

\subsection{The Defender and the Attacker}

In both the co-trained and half-step arms, the final pair-accuracy metric is 1.00 for every defender update in rounds 8 through 12. This metric is the fraction of the last 20 training pairs in an update for which the updated defender favors the chosen response over the rejected response more than the frozen base model does. Over the same last-20-pair window, mean training loss falls from 0.616 in the co-trained arm and 0.624 at half step in round 1 to between 0.02 and 0.06 at rounds 11 and 12. At each check, the updated defenders refuse at most one more of the 56 benign prompts than the base defender (0.018). In the co-trained arm the attack pairs follow the winner mix: the share of hypothetical framing among the 64 attack pairs rises from 0.16 to 0.56 over rounds 1 to 6. The success of the training attacks keeps rising or holds despite the high pair accuracy: 0.57 to 0.65 over iterations 21 to 24 in the co-trained arm, 0.51 to 0.67 in the frozen arm and 0.25 to 0.35 at half step. Over iterations 7 to 12 the success share of the co-trained arm lies below that of the frozen arm by 0.03 to 0.11 at five of six iterations (0.158 to 0.335 against 0.189 to 0.406), a gap that one seed leaves unassigned. Each defender update takes about 12 optimizer steps, so the results do not show how more extensive defender training would affect the outcome.

\subsection{Update Size and Run-to-Run Variation}

Mean KL to the base model was generally low, with a few brief spikes. In the first twelve iterations it was below 3 at every update except iterations 1 and 2 of the frozen arm (192 and 2,562, with largest gradient norms of 693 and 10,178) and iterations 2 and 6 of the co-trained arm (390 and 46, with gradient norms of 1,764 and 156). A later spike reached 979 at iteration 18 in the co-trained arm. The half-step arm stayed at or below 1.2 through iteration 19 (4.2 at iteration 20). The ordering of the arms by spike size matches the ordering by early narrowing (frozen, co-trained, half step), which leaves a head start from large early updates open. The half-step arm nevertheless narrows by 0.295 at iteration 24 with every update small, and the narrowing of the frozen arm continues over iterations 3 to 12 while every update has a mean KL below 3.

Run-to-run variation is large early. The co-trained and frozen arms share a configuration through iteration 2, because the first defender update follows that iteration. They already differ at iteration 2 (ADR 0.915 against 1.074 with disjoint intervals) and in the first attacker update (mean KL 2.23 against 192.06, with largest gradient norms of 6.07 and 692.93). Across the three retained runs the standard deviation of ADR is 0.30 over updates 1 and 2, 0.16 over updates 3 to 8, 0.07 over updates 9 to 20 and 0.03 over updates 21 to 24.

\subsection{Row Cap and Reward}

The update uses up to 512 signal rows per iteration, selected uniformly using the seed from about 1,170 to 1,490 signal rows among the 1,536 attack slots. Selected rows match the whole set: the enrichment of the common opening among selected rows relative to all valid attacks is 0.99 to 1.20 as the mean of each six-iteration window in every arm, and the diversity of selected rows relative to all valid attacks of the same behaviors is 0.98 to 1.01. In the retained runs the family mix of selected rows differs from that of all signal rows by 1.57 percentage points on average (largest 3.63). The reward is the success flag alone, with quality and novelty weights of zero, normalized within each group of 16 attacks and tied to the base model by a KL weight of 0.005. The training objective has no explicit term that rewards diversity among attacks, either within a behavior or across behaviors. Instead, it weights rows by training advantage.

\subsection{Attack Length and Wording}

\begin{figure}[t]
\centering
\definecolor{lenCo}{HTML}{DC2626}
\definecolor{lenFrozen}{HTML}{2563EB}
\definecolor{lenHalf}{HTML}{059669}
\begin{tikzpicture}
\begin{axis}[
    width=0.88\columnwidth, height=5.0cm,
    xmin=0.35, xmax=1.05, ymin=0.5, ymax=3.5,
    xtick={0.4,0.6,0.8,1.0},
    ytick={1,2,3}, yticklabels={Half step,Frozen defender,Co-trained},
    xlabel={attack diversity retention (ADR)},
    tick label style={font=\scriptsize}, label style={font=\scriptsize},
    grid=major, grid style={black!10}, axis line style={black!60},
    tick style={black!60}, clip=false]
  \draw[densely dashed,black!55] (axis cs:1,0.5) -- (axis cs:1,3.5);
  \draw[lenCo,thick] (axis cs:0.493,3.13) -- (axis cs:0.610,3.13);
  \draw[lenCo,thick] (axis cs:0.54,2.87) -- (axis cs:0.83,2.87);
  \draw[lenFrozen,thick] (axis cs:0.465,2.13) -- (axis cs:0.578,2.13);
  \draw[lenFrozen,thick] (axis cs:0.50,1.87) -- (axis cs:0.66,1.87);
  \draw[lenHalf,thick] (axis cs:0.631,1.13) -- (axis cs:0.782,1.13);
  \draw[lenHalf,thick] (axis cs:0.70,0.87) -- (axis cs:0.96,0.87);
  \addplot[only marks,mark=*,mark size=2.3pt,lenCo] coordinates {(0.552,3.13)};
  \addplot[only marks,mark=*,mark size=2.3pt,lenFrozen] coordinates {(0.519,2.13)};
  \addplot[only marks,mark=*,mark size=2.3pt,lenHalf] coordinates {(0.705,1.13)};
  \addplot[only marks,mark=diamond*,mark size=2.5pt,lenCo,mark options={fill=white}] coordinates {(0.68,2.87)};
  \addplot[only marks,mark=diamond*,mark size=2.5pt,lenFrozen,mark options={fill=white}] coordinates {(0.57,1.87)};
  \addplot[only marks,mark=diamond*,mark size=2.5pt,lenHalf,mark options={fill=white}] coordinates {(0.84,0.87)};
\end{axis}
\end{tikzpicture}
\caption{ADR at iteration 24 before length adjustment (filled circles) and after adjustment within length bands (open diamonds). Lines are 95\% behavior-bootstrap intervals. The dashed line marks the untrained reference. Each arm has one seed. The length adjustment is an exploratory sensitivity analysis with thin support, not an estimate of a causal effect of length.}
\label{fig:length-sensitivity}
\end{figure}
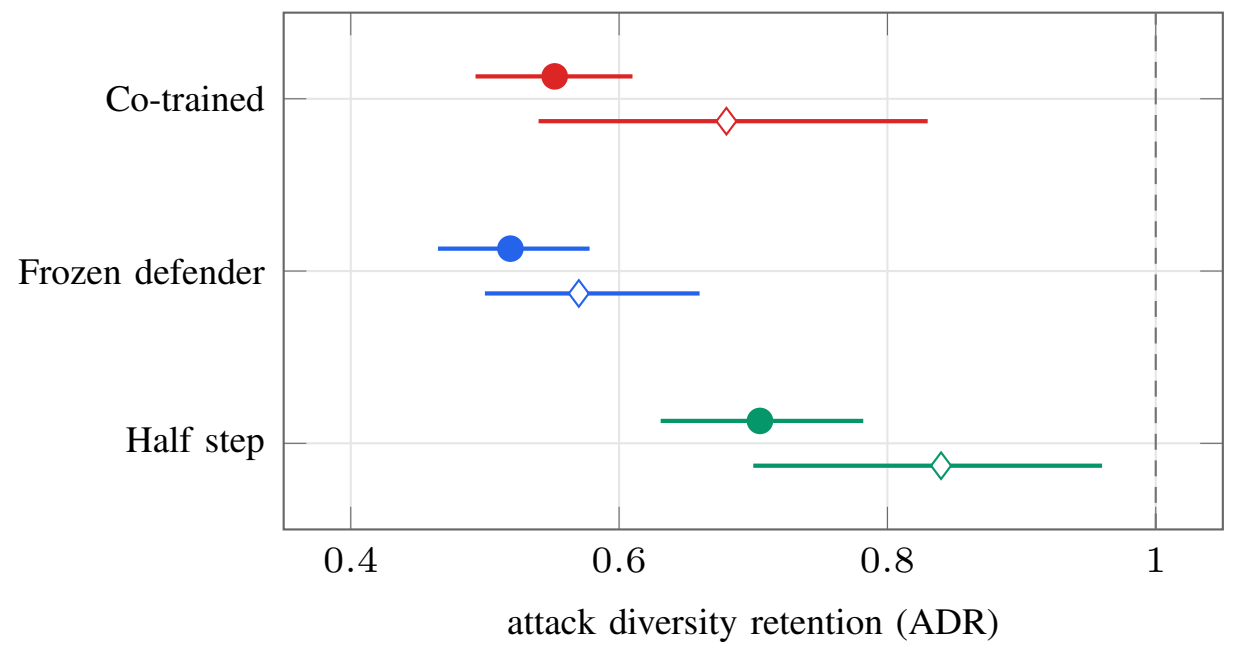

Attacks lengthen as success grows: mean length is 221 characters for the untrained attacker and 311, 418 and 367 characters at iteration 24 (1.4, 1.9 and 1.7 times). ADR after length adjustment is 0.68 (0.54 to 0.83), 0.57 (0.50 to 0.66) and 0.84 (0.70 to 0.96) at iteration 24, compared with unadjusted values of 0.552, 0.519 and 0.705 (Figure~\ref{fig:length-sensitivity}). The adjustment raises ADR by 0.13, 0.05 and 0.14. This increase is roughly half the unadjusted fall at half step, but is not a causal decomposition. On fresh banks at iteration 24, adjusted ADR is 0.54, 0.62 and 0.82, compared with unadjusted values of 0.52, 0.56 and 0.71. In the frozen arm, the adjustment raises ADR by 0.06 to 0.15 at iterations 8 to 10 and by little at iteration 12. The interval for adjusted ADR in the co-trained arm lies below 0.75 on the fresh banks at iterations 20 and 24 (upper ends 0.64 and 0.62) and reaches 0.80 at iteration 16 (estimate 0.69). The adjustment has thin support because most pairs fall in the longest length band.

The wording-based measure does not show a comparable decline. The mean pairwise Jaccard distance between word 3-gram sets of the attacks for one behavior is 0.95 at iteration 1 and stays within 0.98 to 1.04 times that value at every iteration in every arm, so the wording is near saturation, since attacks for one behavior share few 3-grams. What converges is the opening, and the common-opening tracks the dominant attack pattern (Table~\ref{tab:records}). An independent encoder reading is uncollected.

\section{Discussion}
\label{sec:discussion}

\subsection{Where the Narrowing Appears to Originate}

Table~\ref{tab:standing} compares expected outcomes for each of the mechanisms. Four observations place the origin of attack diversity reduction between the judge score and the next attacks generated by the attacker.

First, narrowing proceeds more quickly with a frozen defender. This comparison suggests that defender adaptation (M1) affects the timing of narrowing but is not required for it in this experiment. Second, winners have narrower attacks relative to all other valid attacks at every retained update, starting with the first (M3): the winner dispersion ratio stays between 0.70 and 0.93, the 72 retained updates agree, and the training objective has no explicit term rewarding diversity among attacks for the same behavior. Third, the winner-neighbor advantage is 0.10 to 0.15 and the common-opening advantage is 0.05 to 0.16 in every arm, after a defender update and within a fixed defender alike (M2). Fourth, the row cap and the update transients account for little impact on ADR (M4, M5), and in one retained run the family mix became more balanced while the embedding spread fell (M6).

\begin{table*}[t]
\centering
\caption{Standing of the mechanisms. Ruled out means that the records contradict an expected outcome in Table~\ref{tab:mechanisms}, within the scope of this study. All readings come from one seed per arm and the retained runs.}
\label{tab:standing}
\small
\begin{tabular}{@{}P{0.15\textwidth}P{0.57\textwidth}P{0.22\textwidth}@{}}
\toprule
Mechanism & What the records show & Standing \\
\midrule
M1 Defender adaptation & The frozen arm narrows earlier and ends at the same level (0.519 against 0.552). The contrast is $-0.033$ ($-0.088$ to $0.020$). Final pair accuracy is 1.00 from round 8. The winner-neighbor advantage is unchanged after and between defender updates, and the success share of training attacks holds at 0.57--0.65 against 0.51--0.67. & Defender updates delay narrowing by about six iterations in this setup, but are not required for it. Each update uses about 12 optimizer steps. \\
\addlinespace
M2 Success edge & The winner-neighbor advantage is 0.10--0.15, and the common-opening advantage is 0.05--0.16 in every arm and window, including with a fixed defender. The edge could belong to the judge, the base defender or the content of the attacks. An audit of the judge decisions is uncollected. & Present in every arm. Source remains open. \\
\addlinespace
M3 Reward reinforcement & Winners are narrower than attacks in 72 of 72 retained updates. The ratio ranges from 0.70 to 0.93 starting at iteration 1. The training objective has no explicit diversity reward. In the co-trained arm, mean training advantage is positive for attacks with the common opening first and negative for alternatives later. Likelihoods of fixed attacks are uncollected. & Consistent with every record. The likelihood link remains untested. \\
\addlinespace
M4 Row cap & Enrichment of the common opening among selected rows is 0.99--1.20 and the diversity of selected rows is 0.98--1.01 of all valid attacks. The family mix differs by 1.57 percentage points on average. & Ruled out. \\
\addlinespace
M5 Update size & Mean KL is below 3 at all but a few updates. The half-step arm has only small updates and narrows by 0.295. The arm with the largest transients narrows first. & Ruled out as the main driver. A head start remains open. \\
\addlinespace
M6 Measurement & ADR after length adjustment is higher by 0.05 to 0.14 at iteration 24. Wording-based distance is saturated. In one retained run the family mix became more balanced (concentration index 0.400 to 0.277) while embedding spread fell. Within-behavior distance and opening share agree with ADR. & Length contributes in part. An independent encoder reading is uncollected. \\
\bottomrule
\end{tabular}
\end{table*}

The records show a pattern with two parts. Some attack patterns succeed more often than others, even when the judge and defender stay fixed. This difference could steer training toward the more successful patterns. The success-only training advantage gives winners in mixed-reward groups positive weight. The update could then make those attacks more likely in the next iteration. In the co-trained arm, mean training advantage is positive for attacks with the common opening over iterations 9 to 14 and negative for attacks without it over iterations 17 to 21, as the dominant attack pattern takes over.

This proposed sequence depends on association. A successful outcome depends partly on training, because a policy that repeatedly produces a given attack pattern may execute it more reliably, and the available records cannot determine how much of the success difference reflects training versus the attack pattern itself. The link from training advantage to likelihood change is not measured, because likelihoods of fixed attacks before and after each update have not been collected. Across behavior-update combinations with sufficient data in the retained runs, the correlation between the winner enrichment of a family and the change of the next-iteration share of that family is weak and inconsistent (Spearman 0.147, 0.085 and $-0.009$). Family-level correlations in the arms are higher (0.65, 0.49 and 0.30, with slopes 0.32, 0.26 and 0.07 in the co-trained, frozen and half-step arms) and come from one seed per arm. Analyses of entropy collapse in policy-gradient training tie the change of policy entropy to the covariance between action probability and policy advantage \cite{cui2025entropy}, which supports the plausibility of this interpretation, although this study leaves it untested.

\subsection{Alternatives Ruled out}

Six alternatives conflict with outcomes in the tested arms. M1 would imply weaker narrowing under a frozen defender, but the frozen arm narrows sooner. M4 would imply that selected rows differ from all signal rows, but both the common-opening enrichment (0.99 to 1.20) and the diversity ratio (0.98 to 1.01) stay near 1. M5 would imply that an arm with only small updates retains the diversity of the untrained attacker, but the half-step arm narrows by 0.295. One retained run narrows even as its family shares become more evenly distributed. Narrowing also continues after groups with a single winner contribute little to the positive training advantage. If length were the sole explanation, ADR after length adjustment would remain near 1. Instead, the adjustment raises ADR by only 0.05 to 0.14 at iteration 24.

\subsection{Mechanisms Unaccounted For}

Four aspects remain open. The source of the success edge (M2) may be the preference of the judge, a weakness of the base defender or the content of the attacks. The causal link from training advantage to likelihood (M3) lacks a direct measurement. The effect of defender adaptation (M1) is tested only with about 12 optimizer steps per update. More extensive defender training could yield different results. By adding an independent enconder, the share of the diversity fall that reflects strategy variety can be determined rather than relying on embedding geometry (M6).

\subsection{Delay versus Prevention}

The frozen arm crosses 0.75 six times before the co-trained arm does (iterations 10 and 16), the half-step arm crosses three iterations after the co-trained arm (iteration 19), and the arms end at 0.519, 0.552 and 0.705. Defender updates therefore delay narrowing in this study instead of preventing it, and the halved learning rate delays it further, again short of prevention. At equal success fraction the half-step arm lies between the other arms, so slower progress accompanies the late crossing. The co-trained arm ends at a level like that of the three retained runs at the same iteration (0.552 on the development bank and 0.516 on the fresh bank, against 0.64, 0.60 and 0.62 and 0.63, 0.61 and 0.63).

The timing differences are considered in light of run-to-run variation. At iteration 12, four runs with an updated defender read 0.70, 0.76, 0.70 and 0.891 (mean 0.763, standard deviation 0.090), and the frozen arm (0.496) lies 3.0 standard deviations below that mean. Yet the identical configurations of the first two iterations already differ by 0.16 at iteration 2. The late agreement of the co-trained and frozen arms at iteration 24 (contrast $-0.033$) sits within the between-run standard deviation of 0.03 observed across three retained runs at updates 21 to 24, and the half-step contrast of $+0.153$ lies well above that spread. The half-step arm was flat at 0.69 to 0.71 over iterations 21 to 24, so a continuation toward the other arms remains open.

\subsection{Threshold versus Continuous Reading}

The ADR 0.75 reference separates arms based on continuous readings. The co-trained arm meets the reference on every checkup on the fresh-bank from iteration 16 onwards, whereas the half-step arm keeps an upper interval end of 0.782 at iteration 24, although the cumulative change is $-0.295$ ($-0.370$ to $-0.218$). A threshold reading alone would classify the half-step arm as free of narrowing. The continuous reading shows narrowing that is smaller, delayed and length-sensitive, since the length adjustment removes roughly half of the diversity reduction. The conclusions here follow the continuous readings, and the reference marks timing.

\subsection{Further study of the Open Aspects}

M2 and M3 can coexist, and the present records leave them inseparable. Possible reward interventions include changing training advantages within groups or adding a term that values differences between attacks. Such a term could favor alternative attacks despite the common-opening advantage of 0.05 to 0.16 in success rates. If ADR increased while attack success stayed at the same level, that would show that a reward intervention can increase diversity without reducing success. Likelihoods of fixed attacks before and after each update would measure the link from training advantage to probability directly. A second judge with a human audit would show whether the edge reflects the preference of the judge. Rescoring the same attacks against another defender would show whether it reflects a weakness of the defender. An independent encoder would test the embedding geometry, and further seeds would separate arm effects from run-to-run variation.

\subsection{Implications}

If these interpretations hold, changes to the defender or step size are expected to shift the timing of narrowing, as the frozen and half-step arms show. Changes to the reward are candidates for preventing attack narrowing.

\section{Limitations}
\label{sec:limits}

\begin{enumerate}
    \item \textbf{Seeds and run-to-run variation.} Each arm has only one run, so differences between arms may reflect run-to-run variation as well as the interventions. The co-trained and frozen arms have identical configurations through iteration 2, yet their ADR values differ by 0.16 at that iteration. Their first attacker updates also have mean KL values of 2.23 and 192.06, respectively. Comparisons with the co-trained arm include all effects of each intervention, including changes in success and the signal rows used for training. The intervals describe variation across behaviors within one seed and do not measure variation across seeds. The 72 comparisons from retained runs are correlated observations from three evolving policies, not 72 independent replications. Reporting each run individually follows recommendations for reinforcement learning studies with few runs \cite{henderson2018matters,agarwal2021precipice}.
    \item \textbf{Exploratory status.} The cumulative change at iteration 12 and the slope over iterations 3 to 12, including their arm comparisons, were specified before the experiment. The extension to 24 iterations, mechanism records, equal-progress comparisons, length-adjusted readings and analyses of retained runs are exploratory. None of the readings are corrected for multiple comparisons.
    \item \textbf{Judge.} A single classifier determines whether an attack succeeds, and model judges have known biases \cite{zheng2023judge}. The attacks were not evaluated by a second judge or a human audit. The source of the success edge therefore remains uncertain.
    \item \textbf{Measurement.} ADR compares embedding distances measured by one encoder. Attack length explains part of the observed diversity decline, but the attacks were not evaluated with an independent encoder. The wording-based distance is near its maximum because attacks for the same behavior share few word 3-grams. Strategy-family labels are generated automatically and describe broad categories. Most readings use 25 of the 26 development behaviors. For the half-step trend and its comparison with co-training over iterations 3 to 12, the ADR values are calculated from 24 behaviors.
    \item \textbf{Causal link.} The hypothesized sequence from training advantage to attack likelihood and subsequent attack generation is supported by associations, not a direct causal test. The likelihoods of fixed attacks before and after updates were not measured, and the reward was not changed in a direct intervention. In the retained runs, the association between family enrichment among winners and the change in family share at the next iteration is weak and inconsistent within behavior-and-update groups.
    \item \textbf{Defender update.} In the co-trained and half-step arms, a defender update follows every two attacker iterations. Each defender update uses at most 96 preference pairs and takes about 12 optimizer steps. The conclusions about defender adaptation apply to this schedule and may not extend to stronger or more frequent defender updates.
    \item \textbf{Horizon and progress.} By iteration 24, ASR@8 is 1.0 in every arm, meaning that each development behavior has at least one successful attack among the eight generated attacks. At that point, ASR@8 cannot show whether a larger share of individual attacks succeeds. The exploratory comparison instead groups iterations by the fraction of attacks that succeed and compares ADR at similar success levels. In the 0.4 to 0.5 and 0.5 to 0.6 success ranges, each arm contributes only one to four iterations. Early in the run, attack success increased more slowly in the half-step arm, at roughly half the rate observed in the co-trained arm. Both arms completed the same number of iterations, but their success levels and training signals differed. The comparison therefore includes the learning-rate change and any effects it had on success and the training signals.
    \item \textbf{Scope.} This study uses single-turn attacks, one open-weight attacker model, one open-weight defender model, a success-only reward and one training panel drawn from AdvBench behaviors. The findings do not establish whether the same pattern occurs with multi-turn strategies, deployed systems or adaptive human attackers. The held-out test splits were not scored.
\end{enumerate}

\section{Related Work}
\label{sec:related}

Language-model red teaming includes model-generated test cases \cite{perez2022redteaming}, optimized adversarial suffixes \cite{zou2023universal} and standardized evaluation \cite{mazeika2024harmbench}. Several methods reward diversity directly: curiosity-driven red teaming adds a novelty objective \cite{hong2024curiosity}, Rainbow Teaming applies quality-diversity scenarios \cite{samvelyan2024rainbow}, and Beutel et al.\ reward attacks that differ from past attacks \cite{beutel2024diverse}. These works address the solution, whereas the present study traces the origin of attack diversity reduction.

Prior work also studies attacker training against an adapting defender. Multi-round automatic red teaming alternates improvements to an adversarial model and its target \cite{ge2023mart}. Online self-play reinforcement learning trains attackers and defenders together and reports diverse attacks \cite{liu2025chasing}. The AdvGRPO study reports diversity reduction during attacker--defender co-training with GRPO \cite{bullwinkel2026advgrpo}. The delay of narrowing by defender updates in the present study is in the direction of that report, and the arms nevertheless converge by iteration 24.

Reward-driven narrowing of an output distribution has precedent outside red teaming. Reinforcement learning from human feedback reduces output diversity \cite{kirk2023rlhf}. Policy entropy falls with the covariance between action probability and policy advantage \cite{cui2025entropy}. Reasoning models trained with verifiable rewards reach higher accuracy with few attempts, while base models reach higher pass@k with many attempts \cite{yue2025rlvr}. Repeated learning from narrowed outputs erodes distribution tails \cite{shumailov2024collapse}. This proposed sequence is consistent with these findings and adds loop-specific evidence. Analyses of GRPO identify length bias in per-completion normalization \cite{liu2025drgrpo}. The DAPO work discusses sparse signal from uniform-reward groups \cite{yu2025dapo}. The constant normalizer and the row bounds of the update address these artifacts. The row cap is tested directly in Section~\ref{sec:results}.

\section{Conclusion}
\label{sec:conclusion}

This study traced attacker diversity reduction through the attacker--defender--judge loop with three arms sharing a seed, per-iteration records and 72 retained updates. Narrowing appears to originate between the judge score and the next attacks generated by the attacker. A frozen defender narrows sooner, defender updates delay narrowing by about six iterations, and halving the attacker learning rate delays it further. At iteration 24, ADR is below the untrained reference of 1 in every arm: 0.519 for the frozen-defender arm, 0.552 for the co-trained arm and 0.705 for the half-step arm. Winners are narrower than the attacks from the first iteration. The winner-neighbor advantage is 10 to 15 percentage points, while the common-opening advantage is 5 to 16 percentage points in every arm. The training-row cap and the update transients contribute little to attacker diversity reduction, and attack length accounts for part of the attack diversity decline. The success-only training advantage may make the dominant attack pattern more common.

\bibliographystyle{IEEEtran}
\bibliography{references}

\end{document}